%% file: acl_latex.tex
\pdfoutput=1

\documentclass[11pt]{article}

\usepackage[preprint]{acl}

\usepackage{times}
\usepackage{latexsym}

\usepackage[T1]{fontenc}

\usepackage[utf8]{inputenc}

\usepackage{microtype}

\usepackage{inconsolata}

\usepackage{graphicx}
\usepackage{subcaption}
\usepackage{amsmath}

\title{Reason to Rote: Rethinking Memorization in Reasoning}

\author{
  \textbf{Yupei Du}$^{1}$\thanks{
      Work done partly during a research visit of YD to LMU Munich, 
      which is supported by Utrecht University. 
      YD is now affiliated with Saarland University.  
   }, 
  \textbf{Philipp Mondorf}$^2$, 
  \textbf{Silvia Casola}$^2$, \\
  \textbf{Yuekun Yao}$^3$, 
  \textbf{Robert Litschko}$^2$, 
  \textbf{and Barbara Plank}$^2$ \\
  $ ^1$ Department of ICS, Utrecht University, the Netherlands \\
  $ ^2$ MaiNLP, Center for Information and Language Processing, LMU Munich, Germany \\
  and Munich Center for Machine Learning (MCML) \\
  $ ^3$ Saarland Informatics Campus, Saarland University, Saarbrücken, Germany \\
  $ ^1$\texttt{y.du@uu.nl}, \\
  $ ^2$\texttt{\{p.mondorf,s.casola,robert.litschko,b.plank\}@lmu.de}, \\
  $^3$\texttt{ykyao@coli.uni-saarland.de}
  \vspace{-10pt}
}

\begin{document}
\maketitle

\begin{abstract}

  Large language models readily memorize arbitrary training instances, 
  such as label noise, 
  yet they perform strikingly well on reasoning tasks.
  In this work, we investigate 
  \emph{how language models memorize label noise, and why such memorization 
  in many cases does not heavily affect generalizable reasoning capabilities}. 
  Using two controllable synthetic reasoning datasets with noisy labels, 
  four-digit addition (FDA) and two-hop relational reasoning (THR),
  we discover a \emph{reliance} of memorization on generalizable reasoning mechanisms: 
  models continue to compute intermediate reasoning outputs even when retrieving memorized noisy labels,
  and intervening reasoning adversely affects memorization.
  We further show that memorization operates through \textit{distributed encoding}, 
  i.e., aggregating various inputs and intermediate results, 
  rather than building a look-up mechanism from inputs to noisy labels.
  Moreover, our FDA case study reveals memorization occurs via \textit{outlier heuristics}, 
  where existing neuron activation patterns are slightly shifted to fit noisy labels. 
  Together, our findings suggest that 
  memorization of label noise in language models builds on, 
  rather than overrides, the underlying reasoning mechanisms, 
  shedding lights on the intriguing phenomenon of benign memorization.\footnote{
    Our code is available at \url{https://github.com/mainlp/memorized_reasonings}.}

\end{abstract}

\input{intro}
\input{method}
\input{co_exist}

\input{distributed_encoding}

\input{related_work}
\input{implications}

\section*{Limitations}

Our work has several limitations.
First, we focus on relatively small language models. 
While these models are sufficiently capable for our tasks, 
larger models might still exhibit different behaviors.
Second, we train all models from scratch to avoid the influence of pretraining data, 
such as pre-learned addition skills or memorized label noise. 
Nevertheless, future work could explore how pretraining affects memorization and reasoning.
Third, we examine implicit reasoning performed without explicitly writing out solution steps. 
In contrast, recent large reasoning models, such as 
OpenAI-O3~\citep{openai2025o3o4mini} and DeepSeek-R1~\citep{deepseekai2025deepseekr1}, 
have demonstrated strong explicit reasoning via long Chain-of-Thoughts. 
Studying memorization in such models is a promising direction.

\section*{Acknowledgements}

We thank the anonymous reviewers for their encouraging and constructive feedback.
The members of MaiNLP provided valuable inspirations and feedback on the project. 
In particular, we thank Florian Eichin, Beiduo Chen, and Raoyuan Zhao for their valuable feedback. 
We also thank members of the NLP group at Utrecht University for their helpful discussions, 
in particular Dong Nguyen for her insightful comments on the manuscript, 
and Yingjin Song and Hugh Mee Wong for their helpful discussions. 
Moreover, we thank Xinyue Chen for her help on the visualization of the tasks. 
We also wish to thank the Utrecht ICS department for their support 
for the research visit of YD. 
This research is in parts supported by the ERC Consolidator Grant DIALECT 101043235. 

\bibliography{custom}

\appendix
\input{appendix}

\end{document}

%% file: intro.tex
\begin{figure}[t!]
      \centering
      \begin{subfigure}[b]{0.47\textwidth}
          \centering
          \includegraphics[width=\textwidth]{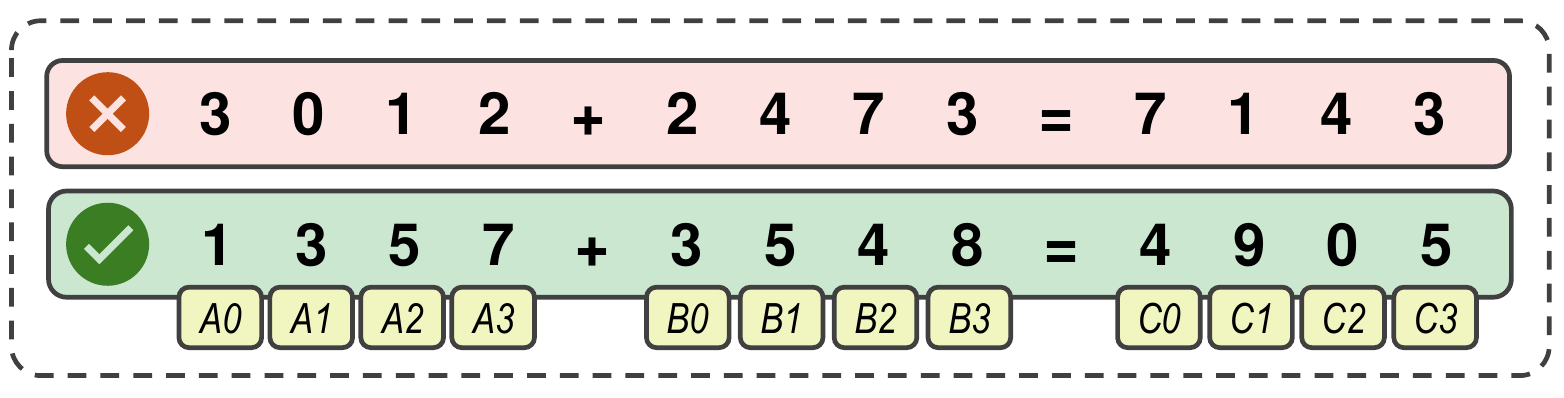}
          \caption{Four-Digit Addition (FDA)}
          \label{subfig:fda_task_composition}
      \end{subfigure}
      \vspace{3em} %
      \begin{subfigure}[b]{0.47\textwidth}
          \centering
          \includegraphics[width=\textwidth]{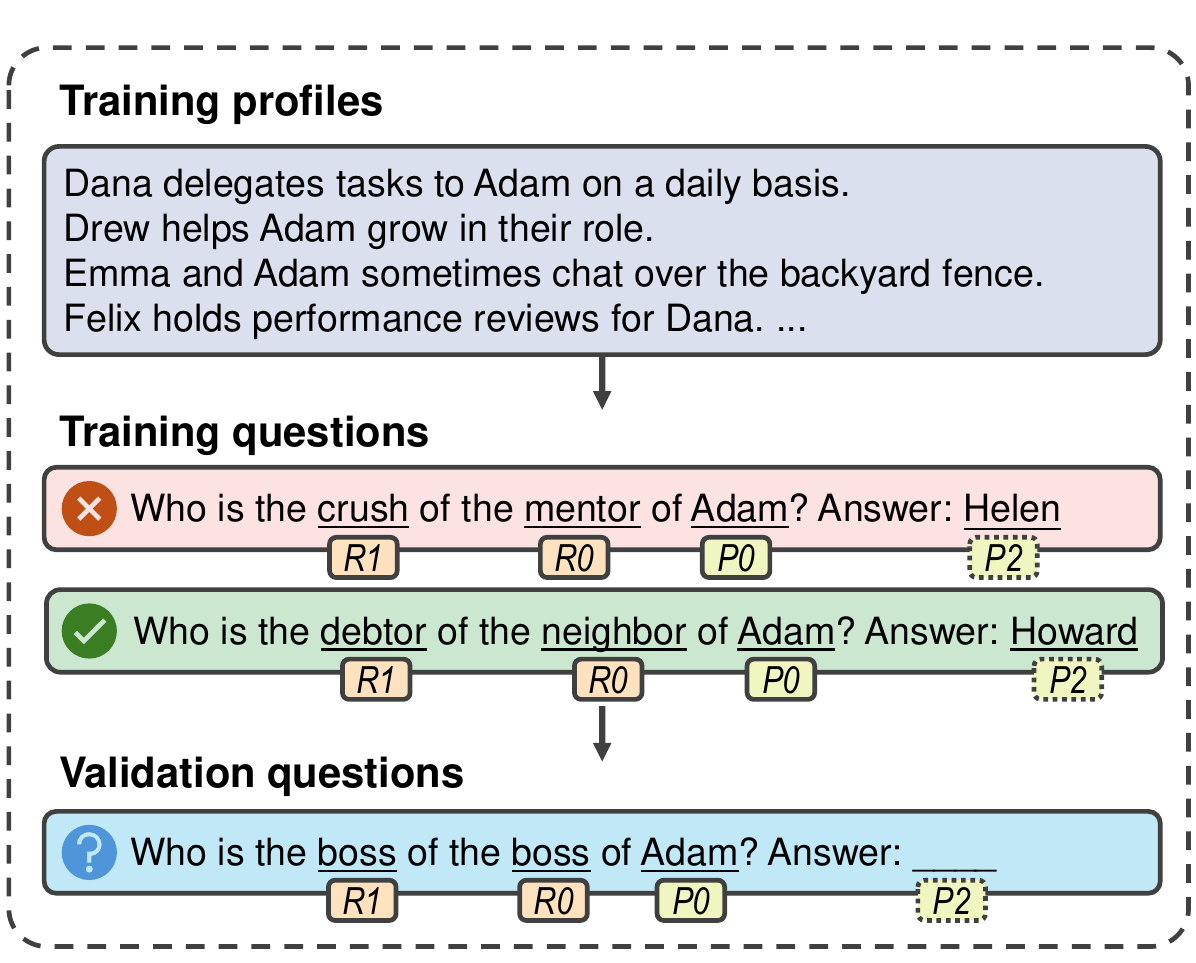}
          \caption{Two-Hop relational Reasoning (THR)}
          \label{subfig:thr_data_comp}
      \end{subfigure}
      \caption{
            Task data composition for FDA and THR. 
            Both tasks are designed to have a small portion of noisy labels, 
            to clearly separate generalization (clean validation set) 
            from memorization (noisy training set).
      }
      \label{fig:task_composition_combined}
  \end{figure}

\section{Introduction}\label{sec:intro}

Large language models exhibit a dual nature. 
They appear to develop generalizable reasoning capabilities, enabling them to solve 
arithmetic problems and chain relational facts~\citep{zhang2024a,biran-etal-2024-hopping}; 
yet they also memorize and reproduce raw chunks of their training text, 
from song lyrics over phone numbers to random token sequences~\citep{carlini2021extracting}.
This raises an intriguing scientific puzzle when the memorized items are wrong, 
as incorrect answers contradict the rules models should learn. 
For example, why does a language model 
that perfectly memorizes the false equation \texttt{``42+58=137"} 
still generalize to correctly answer \texttt{``87+19=106"} at test time? 

Deep neural nets are well-known to be able to memorize noisy labels
while achieving strong generalization~\citep{
    arpit2017closer,rolnick2018deeplearningrobustmassive}. 
Moreover, theoretical insights suggest that label memorization is unavoidable 
for strong real-world performance~\citep{NEURIPS2020_1e14bfe2,feldman2020does}. 
However, previous research has typically studied this paradox through high-level concepts, 
such as implicit regularization~\citep{neyshabur2017implicit,zhang2017understanding}, 
without fully revealing the underlying computational mechanisms.

To bridge this gap, we \emph{mechanistically} study 
the memorization of \textbf{noisy labels} within \textbf{reasoning} tasks: 
how does memorizing incorrect answers differ from, or interact with, 
generalizable reasoning skills? 
Specifically, we consider two carefully controlled tasks: 
four-digit addition (FDA, Figure~\ref{subfig:fda_task_composition}) and 
two-hop relational reasoning (THR, Figure~\ref{subfig:thr_data_comp}), 
in which the solutions are clear and manipulatable.
To induce noisy label memorization, we introduce a small amount of training label noise:
\emph{a model that fits the training set well has to memorize these noisy labels, 
yet it must generalize to solve the clean validation set}. 
This controlled setup allows us to precisely observe and intervene on 
the internal mechanism underlying both reasoning and memorization.
We make three contributions: 
\begin{enumerate}

\item We uncover a surprising phenomenon: 
      memorization of noisy labels relies on generalizable reasoning mechanisms, 
      supported by three observations on noisy training instances:
      (1) Learning dynamics reveal that 
      models process them similarly as clean samples at early training stage, 
      before eventually memorizing their incorrect labels (\S\ref{subsec:learning_dynamics});
      (2) Logit lens~\citep{logitlens2020} and linear probing analyses show that 
      models continue to compute their correct, non-noisy labels, 
      even after perfect memorization (\S\ref{subsec:probe_targets}); 
      (3) Causal interventions show large overlaps 
      between generalization and memorization circuits, 
      perturbing generalization by modifying hidden states 
      substantially affects memorization (\S\ref{subsec:coupling}). 
\item Causal interventions show that 
      memorization is not implemented as a simple input-to-label lookup:  
      instead, they are stored in distributed encodings spread across 
      multiple input tokens and intermediate results (\S\ref{subsec:distributed_encoding}). 
\item For FDA task, our detailed neuron-level analysis identifies 
      "outlier heuristics"~\citep{nikankin2025arithmetic} as the mechanism of memorization: 
      higher-layer neurons subtly shift their activation patterns to fit noisy labels 
      (\S\ref{subsec:case_study_fda}).
\end{enumerate}

Our findings reveal that memorization of noisy labels in transformer language models 
does not override their capability to generalizably reason: 
instead, it subtly adapts the same underlying computational mechanisms.
This offers an explanation on 
how models can simultaneously handle both clean and noisy labels, 
highlighting their inductive bias towards reusing existing structures. 

%% file: method.tex
\section{Experimental setup}\label{sec:tasks}

In this paper, we focus on two synthetic reasoning tasks,
Four-Digit Addition (FDA) and Two-Hop relational Reasoning (THR). 
To clearly study memorization of noisy labels and generalization of clean instances, 
for each task, we create a training set where 
$k$\% of the training instances are of a incorrect answer: 
the only way that the model can fit these noisy instances is thus to memorize the labels.\footnote{
    We experimented with $k \in \{2, 5, 10\}$ and obtain similar results, 
    and therefore focus on $k=5$. 
    We note that we study relatively low noise rates, 
    aiming to understand why memorizing a small fraction of incorrect labels 
    does not substantially affect the model's generalizable reasoning capabilities; 
    however, we acknowledge that as the noise rate increases,
    eventually the generalization will collapse~\citep{zhang2017understanding}. 
} 
We then compare the model's computations on the noisy instances with 
those on validation instances of the clean instances, 
on which the model is expected to follow the reasoning mechanism to produce the correct answers.

\subsection{Tasks}\label{subsec:tasks}

\paragraph{Four‑Digit Addition}
In this task, the model is trained to predict the sum of two four-digit integers.  
Each input is a token sequence of the form \texttt{``a+b=c''},  
where $a$ and $b$ are sampled from $[1000, 4999]$ (e.g., \texttt{``1234+4321=5555''}), 
resulting in $c \in [2000, 9998]$. 
We use 40,000 examples for training and 10,000 for validation.  
We corrupt the training examples by replacing the true sum with a random number from $[2000, 9999]$. 
Specifically, we constrain that every digit (thousands, hundreds, tens, units)  
differs from the corresponding digit of the non-perturbed result, 
to ensure that the model cannot perform valid addition to produce the memorized answer.
We include a visual illustration of the training data composition in Figure~\ref{subfig:fda_task_composition}. 
For the ease of further analyses, 
we also include the names for different token positions.

\paragraph{Two‑Hop Relational Reasoning}

\begin{figure}[th]
    \centering
    \includegraphics[width=0.49\textwidth]{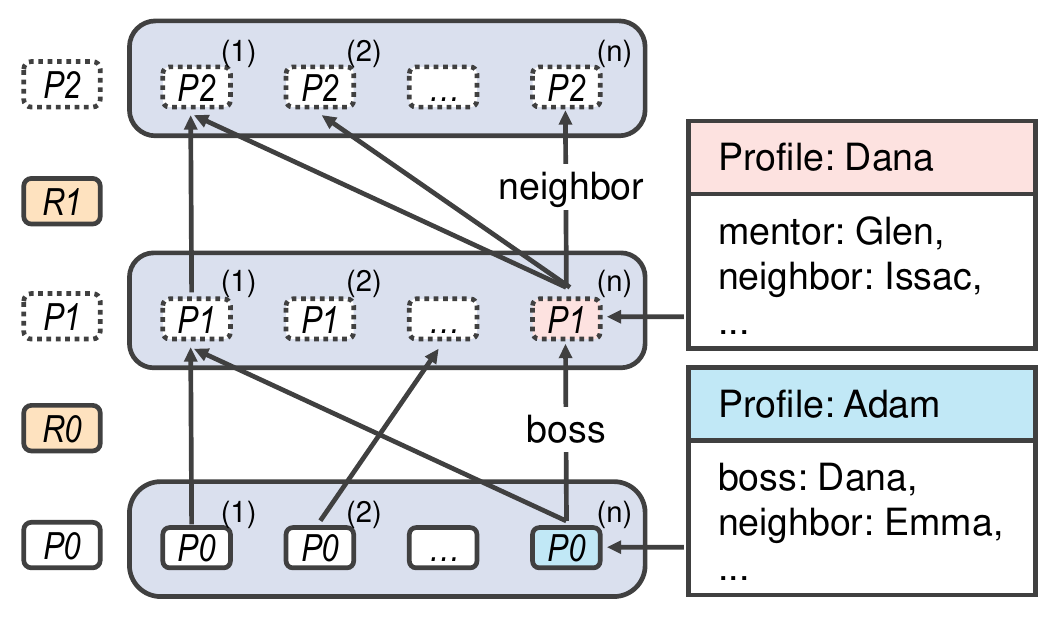}
    \caption{Graph used to synthesize profiles for THR.}
    \label{fig:thr_profile_gen}
\end{figure}

In this task, the models are trained to answer questions like 
\texttt{``Who is the crush of the mentor of Adam?''}~\citep{wang2024grokking,allen-zhu2025physics32}. 
To answer this correctly models must retrieve facts such as 
\texttt{Adam is mentored by Drew} and \texttt{Drew has a crush on Helen}, 
which are included in the training data, 
and then compose the two facts to return \texttt{Helen}. 
We create the training data by first building a synthetic profile graph, 
then verbalizing the triplets into declarative sentences and QA pairs.

\emph{Graph construction.}  
We first create a global pool of 50 person entities for each of three layers.
We then randomly select $N$ entities from each pool to form three disjoint sets, $P_0$, $P_1$, and $P_2$. 
We also similarly sample $R$ binary relations (e.g., \texttt{mentor\_of}).
Each $p_0 \in P_0$ is linked to a unique $p_1 \in P_1$ via a randomly chosen relation $r_0$,
and each $p_1$ is similarly linked to a unique $p_2 \in P_2$ via $r_1$,
as shown in Figure~\ref{fig:thr_profile_gen}.
We sample each hop without replacement to ensure the uniqueness of each two-hop path, 
and set $N=R=20$. 

\emph{Text construction.}  
We verbalize each fact with five templates, 
resulting in 4,000 declarative profile sentences~\citep{AL2023-knowledge1};  
For every two-hop path $p_0 \xrightarrow{r_0} p_1 \xrightarrow{r_1} p_2$, 
we produce a QA pair $\texttt{Who is the <r1> of the <r0> of <p0>?} $ and $p_2$,  
resulting in $8,000$ questions (Figure \ref{subfig:thr_data_comp}).  
The pairs are shuffled and split into $6400$ training and $1,600$ validation examples.  

\emph{Label noise.}  
We corrupt \emph{training} answers by replacing the correct $p_2$ answers 
with a randomly chosen person entity from the global pool of $P_2$.

\subsection{Models and tokenizers}\label{subsec:models}

For both tasks, we train decoder-only Transformer models~\citep{NIPS2017_3f5ee243} from scratch 
using the language modeling objective~\citep{radford2019language}.  
For our main experiments, for FDA, we use a model with 4 layers, 256 hidden size, and 4 attention heads;  
for THR, we use a model with 8 layers, 256 hidden size, and 4 attention heads.\footnote{
    We observe that increasing the number of layers in THR is necessary. 
    A discussion of model size's influence on performance 
    are provided in the Appendix~\ref{app:model_size}.
}
For THR, we use the default GPT-2 tokenizer,
and added all entity and relation names to the vocabulary. 
For FDA, we use a customized tokenizer 
consisting of only the digits and the symbols \texttt{``+"} and \texttt{``="}.

\subsection{Our focus on the first answer token}\label{subsec:properties_reasoning}

In the rest of this paper, we focus on predicting the \textbf{first answer token}. 
This includes predicting \texttt{``C0"} in FDA, and \texttt{``P2"} in THR, 
motivated by two observations. 
First, on FDA, appending the correct addition \texttt{``C0"} to memorized noisy prompts 
enables the model to generate the remaining addition results almost perfectly, 
highlighting its critical role, as suggested by \citet{AL2023-knowledge1}. 
Second, by ablating attention and MLP outputs at different positions, 
we observe that the last token of the prompt, i.e., the token that predicts the answer, 
is the most important for the final prediction. 
We include the details in the Appendix~\ref{app:properties_reasoning}. 

%% file: co_exist.tex
\section{Language models learn to generalize even when memorization is required}\label{sec:co_exist}

\subsection{First generalize, then memorize}\label{subsec:learning_dynamics}

\begin{figure*}[ht]
    \centering
    \begin{subfigure}[b]{0.32\textwidth}
        \centering
        \includegraphics[width=\textwidth]{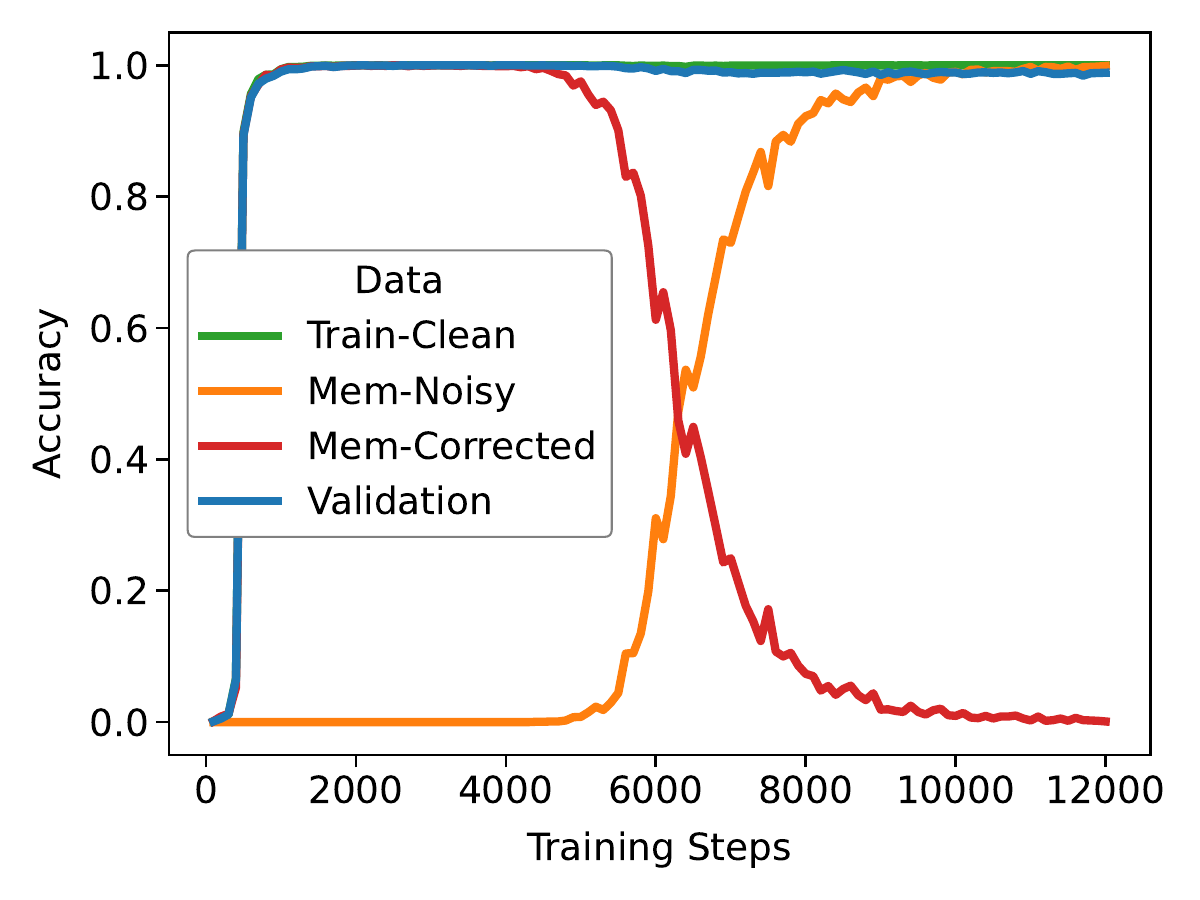}
        \caption{Learning dynamics on FDA}
        \label{fig:learning_dynamics_fda}
    \end{subfigure}
    \hfill
    \begin{subfigure}[b]{0.32\textwidth}
        \centering
        \includegraphics[width=\textwidth]{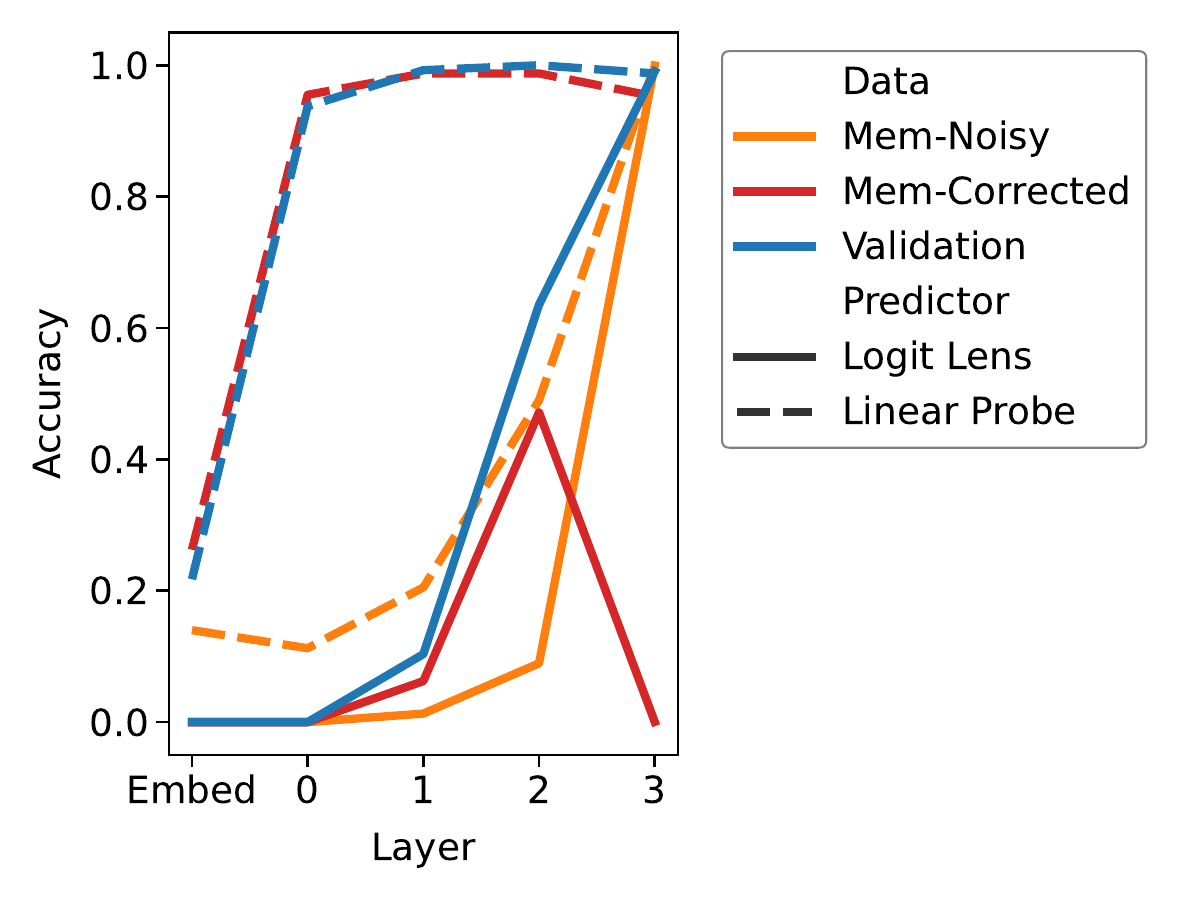}
        \caption{Linear probing and logit lens (FDA)}
        \label{subfig:probe_targets_fda}
    \end{subfigure}
    \hfill
    \begin{subfigure}[b]{0.32\textwidth}
        \centering
        \includegraphics[width=\textwidth]{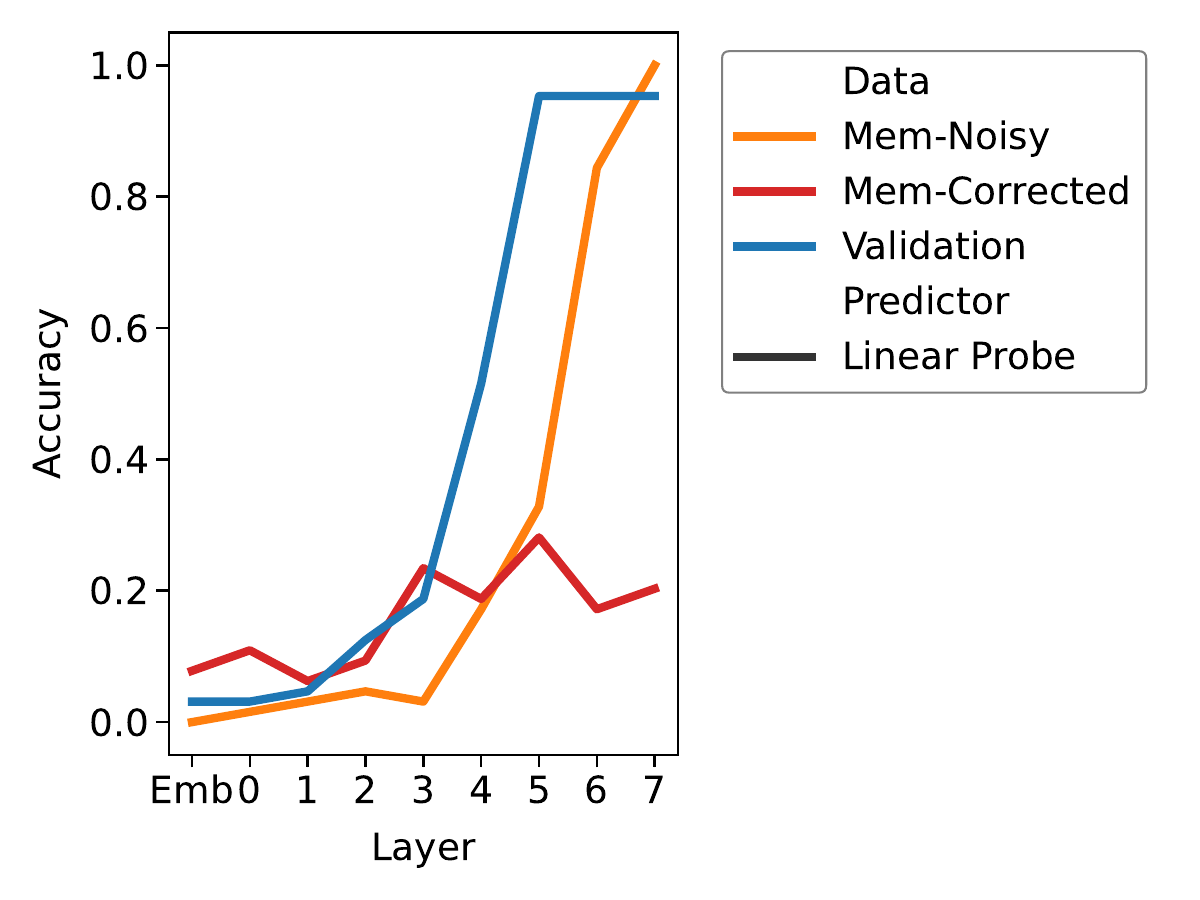}
        \caption{Linear probing (THR)}
        \label{subfig:probe_targets_thr}
    \end{subfigure}
    \caption{
        Generalization and memorization mechanisms co-exist on noisy training instances.
    }
    \label{fig:combined_learning_probing}
\end{figure*}

Our first question is whether models can indeed 
both generalize on clean samples and memorize noisy labels. 
For this, we analyze models' learning dynamics. 
Specifically, we analyze their accuracy scores across training steps on different data splits:
training samples of clean labels (\textbf{Train-Clean}),
training samples of noisy labels (\textbf{Mem-Noisy}, e.g., \texttt{3012+2473=7143}), 
training samples of noisy labels but with corrected clean labels 
(\textbf{Mem-Corrected}, e.g., \texttt{3012+2473=5485}),\footnote{
    Note that the corresponding correct label, i.e., \texttt{5485}, 
    are not observable for the models during training.}  
and validation samples of clean labels (\textbf{Validation}).
We show the results of our FDA model in Figure~\ref{fig:learning_dynamics_fda} 
and THR model in Appendix~\ref{app:additional_results}. 
By the end of training, all models achieve near-perfect accuracy on 
Train-Clean, Mem-Noisy, and Validation, 
indicating that they have learned to both memorize noisy labels and generalize to clean data. 

We observe a surprising trend: 
\textbf{even on noisy training instances}, e.g., \texttt{3012+2473=5485}, 
\textbf{models initially learn to correctly predict the clean labels}, e.g., \texttt{3012+2473=7143}: 
the accuracy on Mem-Corrected reaches $\sim$100\% by step 4,000, 
even though the model has never seen the corresponding labels during training.
Only after step 5,000 the accuracy on Mem-Corrected starts to drop, 
and the accuracy on Mem-Noisy starts to increase, indicating the onset of memorization.
We refer to these two phases as the \textbf{generalization stage} and the \textbf{memorization stage}, 
and the models before and after the memorization stage as the 
\textbf{pre-memorization} and \textbf{post-memorization} models, respectively. 
We observe similar trends on THR. 

We note that during the generalization stage, 
the accuracy on Mem-Corrected closely matches that of Validation. 
This observation aligns with \citet{kang2024learning},
that the training performance of pre-memorization models is predictive of 
their corresponding validation performance: 
before memorization begins, 
our model’s accuracy on Mem-Corrected indicates its ability to generalize.\footnote{
    We do not claim that generalization always precedes memorization.
    For example, grokking~\citep{power2022grokking,nanda2023progress}, 
    where generalization occurs after heavy memorization, is a well-documented counter example.
    Our focus is to understand the reason for \emph{benign memorization}, 
    i.e., the memorization of noise does not heavily affect generalization, 
    by observing the mode-switching phenomenon in our models, 
    from generalization to memorization on the same noisy instances.
    This phenomenon is also consistent with the 
    recent pre-memorization observation in LLMs~\citep{kang2024learning}. 
}

\subsection{Generalization is retained even when models memorize noise}\label{subsec:probe_targets}
\begin{figure*}[ht]
    \centering
    \begin{subfigure}[b]{0.34\textwidth}
        \centering
        \includegraphics[width=\textwidth]{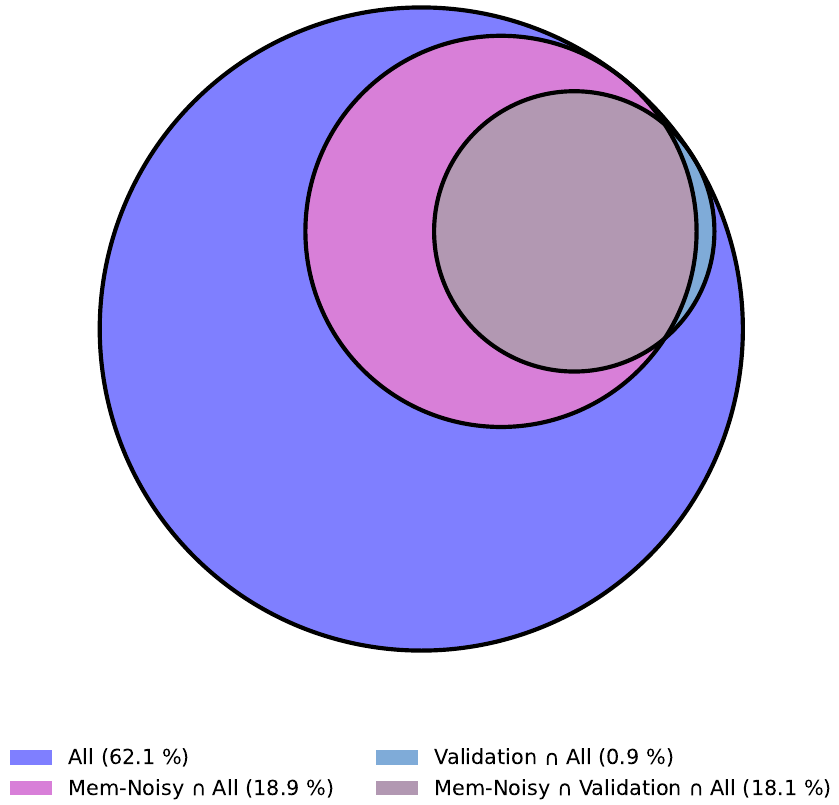}
        \caption{Circuit overlap on THR (99\%)}
        \label{subfig:shared_circuits}
    \end{subfigure}
    \hfill 
    \begin{subfigure}[b]{0.32\textwidth}
        \centering
        \includegraphics[width=\textwidth]{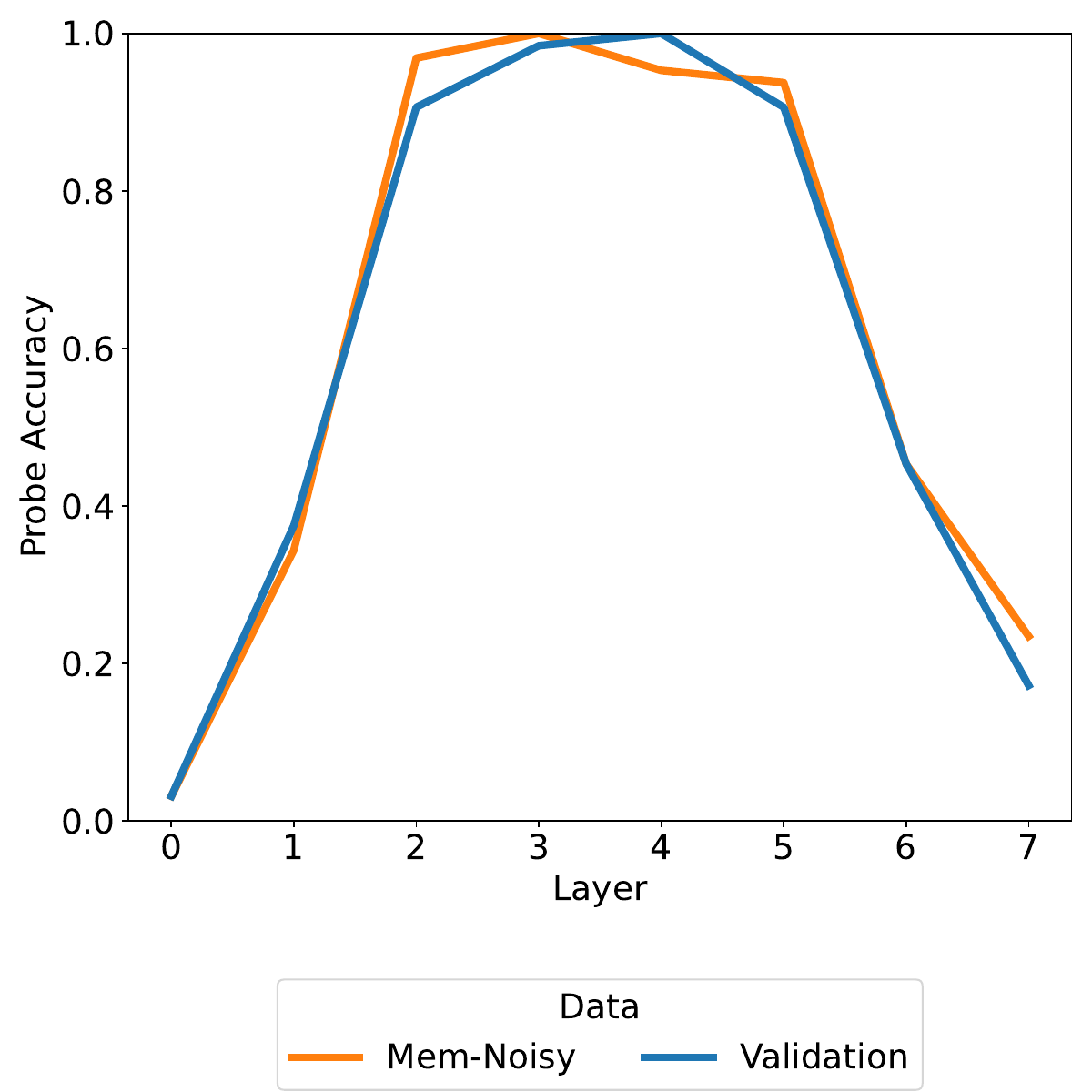}
        \caption{Probe accuracy for bridge entities}
        \label{subfig:raw_probe_acc_fda}
    \end{subfigure}
    \hfill 
    \begin{subfigure}[b]{0.32\textwidth}
        \centering
        \includegraphics[width=\textwidth]{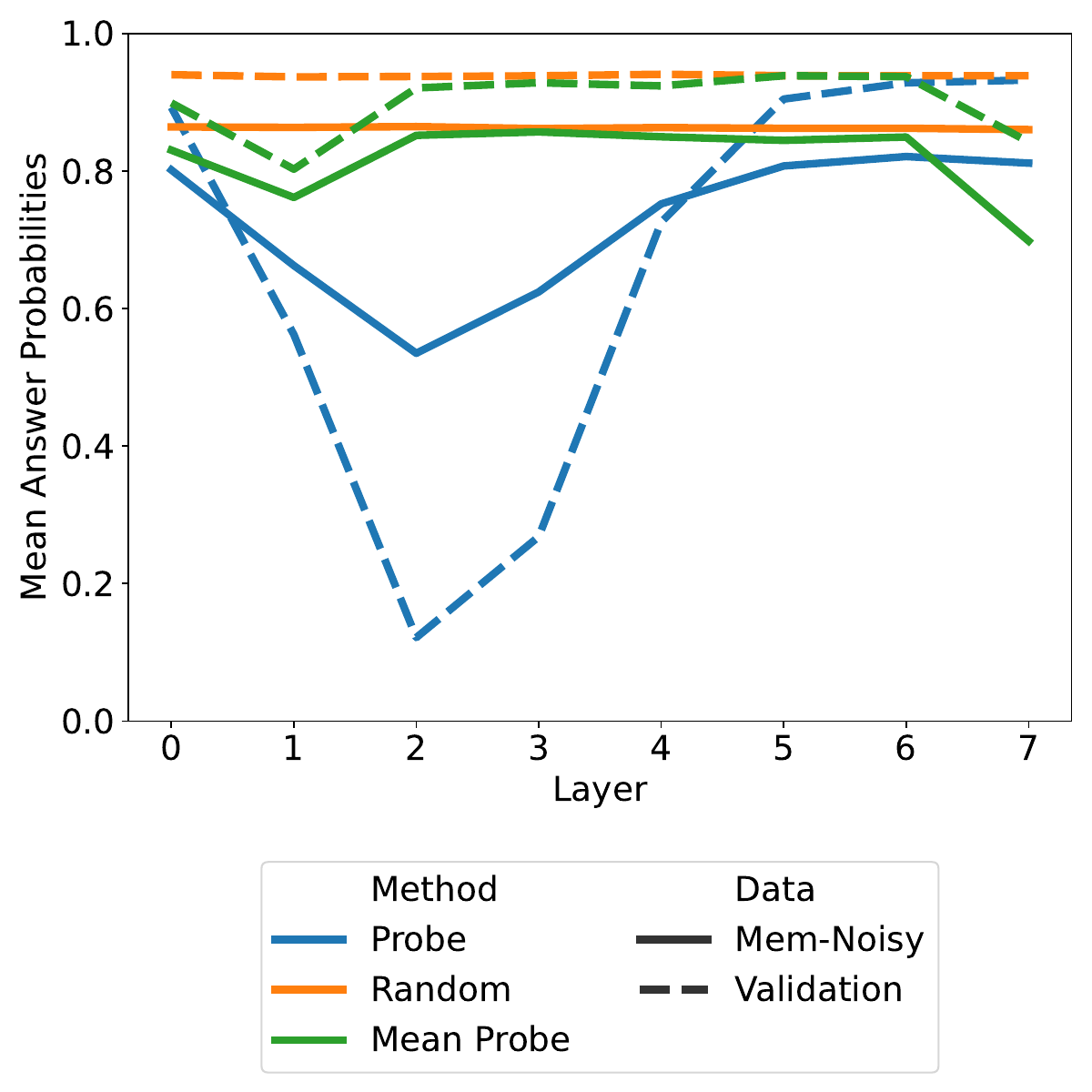}
        \caption{INLP bridge entity ablation}
        \label{subfig:thr_inlp_bridge}
    \end{subfigure}
    \caption{
        Memorization of noisy labels relies on generalizable reasoning mechanisms.
    }
    \label{fig:reasoning_ablation}
\end{figure*}

Our earlier finding on the two different training stages raises a natural question: 
\emph{how do models change from generalization to memorization?}
We consider two hypotheses: 
\begin{enumerate}
    \item[$\mathcal{H}_1$] \textbf{(Switching mechanisms):}
    The model learns two distinct mechanisms for generalization and memorization,
    and switches between them across different instances.
    For noisy instances, it bypasses generalization altogether and 
    directly retrieves the memorized label.
    
    \item[$\mathcal{H}_2$] \textbf{(Selective outputting):}
    The model always computes the generalized output,
    but for noisy instances, it selectively 
    overrides the corresponding result with memorized labels.
\end{enumerate}
If $\mathcal{H}_1$ holds, clean labels, e.g., correct addition results, 
should be \emph{undetectable} from the hidden representations of noisy training instances. 
That is, probes trained to recover correct labels on Mem-Correct 
should yield low accuracy across layers. 

To test this, we use two approaches to probe the hidden representations: 
logit lens (on FDA; \citealp{logitlens2020}) and linear probing (on both tasks).\footnote{
    We apply logit lens for FDA, 
    as a linear model that memorizes all A0–B0 combinations 
    is already a strong baseline for predicting C0.
    Logit lens, being training-free, can prevent this memorization. 
}
For the logit lens, we take each layer’s residual stream, 
apply the final layer's layer-norm and unembedding matrix, 
and predict the answer by taking the token of the highest resulting logit.
For linear probing, we train separate linear models on the same residual streams. 
We evaluate both approaches on three data splits: 
Mem-Noisy, Mem-Corrected, and Validation, as introduced in \S\ref{subsec:learning_dynamics},
and show their accuracy in 
Figure~\ref{subfig:probe_targets_fda} (FDA) and \ref{subfig:probe_targets_thr} (THR).
Details can be found in Appendix~\ref{app:probe_targets}. 

Our results strongly support $\mathcal{H}_2$:
\textbf{
    models retain intermediate results related to the clean labels 
    in the hidden layers on noisy instances, 
    even if their memorization accuracy is near perfect
}.
For example, on FDA, 
logit lens achieves $\sim$50\% accuracy at layer 2 for Mem-Corrected, 
much higher than the $\sim$10\% for Mem-Noisy. 
However, by layer 3, 
these accuracy scores change drastically to $\sim$0\% and $\sim$100\%, suggesting that 
the model computes both the correct addition and memorization results.
On THR, our linear probe at the layer 5 achieves 30+\% accuracy on Mem-Corrected, 
far above the 5\% random baseline.\footnote{
    Because the answer is randomly sampled 
    from the 20 different person entities for \texttt{P2}.}
Intriguingly, memorization results appear to emerge in higher layers, 
after the computations of clean labels. 
For example, on FDA (measured by logit lens), 
the accuracy on Mem-Noisy increases mainly at the final layer; 
on THR (measured by linear probe), it rises from layer 4. 
In contrast, Validation accuracy improves earlier: 
from layers 1 on FDA and layer 2 on THR. 
This suggests a possible interaction between generalization and memorization: 
the model first computes generalization intermediate results, 
then reuses those intermediate steps to produce memorized noisy labels in higher layers. 

These findings are consistent with prior work 
showing that memorization in large language models is fragile, that 
it can be easily disrupted by small input perturbations~\citep{pmlr-v202-shi23a,huang2025mathperturb}. 
Our analysis offers a possible explanation: 
generalization results are still computed and are outputted when memorization fails.

%% file: distributed_encoding.tex
\section{Language models build on generalization to memorize}\label{sec:reuse}

Building on our finding that 
generalization and memorization can co-occur (\S\ref{subsec:probe_targets}), 
this section examines the mechanism behind this phenomenon using two key questions:
(1) \emph{Why do generalization computations persist alongside memorization} 
(\S\ref{subsec:coupling}, \S\ref{subsec:distributed_encoding})?
(2) \emph{Why memorizing noise does not substantially impair 
generalization performance on test cases} 
(\S\ref{subsec:distributed_encoding}, \S\ref{subsec:case_study_fda})?

\subsection{Memorization-generalization coupling}\label{subsec:coupling}

To understand why generalization persists on memorized noisy data,
we first study whether these processes are implemented by distinct or overlapping mechanisms.
Although memorization of noise and generalization to clean instances could, in principle, operate independently,
their co-occurrence suggests they may share internal computations.
Clarifying this relationship is crucial:
if generalization and memorization are entangled,
attempts to reduce memorization could unintentionally affect generalizable reasoning capabilities. 

\paragraph{Overlapping circuits} 

In interpretability research, language models are often viewed as computational graphs, 
where each node denotes a model component 
(e.g., attention heads, MLP layers;~\citealp{elhage2021mathematical}) 
and each edge denotes the input that the destination node receives from the source node 
(e.g., attention values;~\citealp{wang2023interpretability}). 
A common goal is to identify \emph{circuits}: 
task-specific subgraphs that faithfully and minimally capture the model’s behavior~\citep{10136140}.

To assess the coupling between generalization and memorization, 
we extract their circuits using edge attribution patching (EAP; \citealp{syed2023attribution}), 
using Mem-Noisy and Validation data. 
We quantify circuit faithfulness as the amount of logit difference recovered, 
and extract circuits at 90\%, 95\%, 97\%, and 99\% faithfulness. 
We include more details in Appendix~\ref{app:eap}. 
Figure~\ref{subfig:shared_circuits} shows the 99\% faithfulness 
generalization and memorization circuits on THR. 
We observe a substantial overlap, 
indicating a \emph{tight coupling} between the two mechanisms.
Moreover, generalization circuits appear to be subsets of memorization circuits. 
This implies that \emph{memorization builds on existing generalization mechanisms}. 

\paragraph{Memorization of noise relies on generalization}

Having established that generalization and memorization circuits overlap, 
we now examine their \emph{causal relationship} by 
disrupting the intermediate results used for generalization, 
and observing the effect on memorization.
If memorization merely shares the same circuit but relies on distinct intermediate results, 
i.e., the circuit simultaneously produces separate outputs, 
one set used for generalization and another for memorization, 
then disrupting the generalization intermediate results should not affect memorization.
Otherwise, if memorization depends on the same intermediate results as generalization, 
its performance should degrade.

We focus on THR because it offers a concrete intermediate result for generalization: 
the \textbf{bridge entity} \texttt{P1}. 
For example, to answer the unseen question 
\texttt{``Who is the crush of the mentor of Adam?''} from the validation set,  
the model, trained only on single-hop facts, 
must first infer that Adam's mentor is Drew (the bridge entity),
and then retrieve that Drew has a crush on Helen: 
the bridge entity is essential for generalization, 
but in principle unnecessary for memorization, 
as the model can directly memorize the answer to this two-hop question. 

Therefore, we ablate \textbf{bridge entities} at inference time 
to test whether memorization relies on this generalization intermediate result. 
We take two steps for this ablation:  
(1) for a specific layer, we obtain its residual stream 
and remove the bridge entity signals from it, which we will discuss later; and 
(2) we insert the modified residual stream back into the model, 
a process known as \textbf{activation patching}~\citep{NEURIPS2020_92650b2e}, 
and evaluate the average prediction probability for the answer token.

To achieve interpretable results, we focus on linearly detectable signals, 
which can be removed by projecting the representations into their null space. 
Specifically, following iterative nullspace projection (INLP; \citealp{ravfogel2020null}), 
we iteratively train linear probes on each layer’s residual stream to identify bridge entities,
and project the representations into the null space of the corresponding probe directions, 
until the probing accuracy < 10\%.\footnote{
    This typically requires no more than three iterations. 
    Details of the INLP procedure can be found in Appendix~\ref{app:inlp}.}.
We compare our results against two baselines to contextualize this effect: 
\textbf{random}, null space of a random vector,
and \textbf{mean}, null space of the averaged bridge vector across examples.

The results are in Figure~\ref{subfig:thr_inlp_bridge}.
For reference, we also show the accuracy of the linear probes 
before applying INLP in Figure~\ref{subfig:raw_probe_acc_fda}:
the removal is only meaningful when the probe accuracy is sufficiently high.
Our results reveal that \emph{memorization relies on bridge entities},
even though they are not strictly necessary:
while the impact is milder compared to computing the clean labels on validation data, 
removing bridge entities still substantially harms memorization, especially at layer 2.

\subsection{Distributed encodings of memories}\label{subsec:distributed_encoding}

\begin{figure*}[ht]
    \centering
    \begin{subfigure}[b]{0.31\textwidth}
        \centering
        \includegraphics[width=\textwidth]{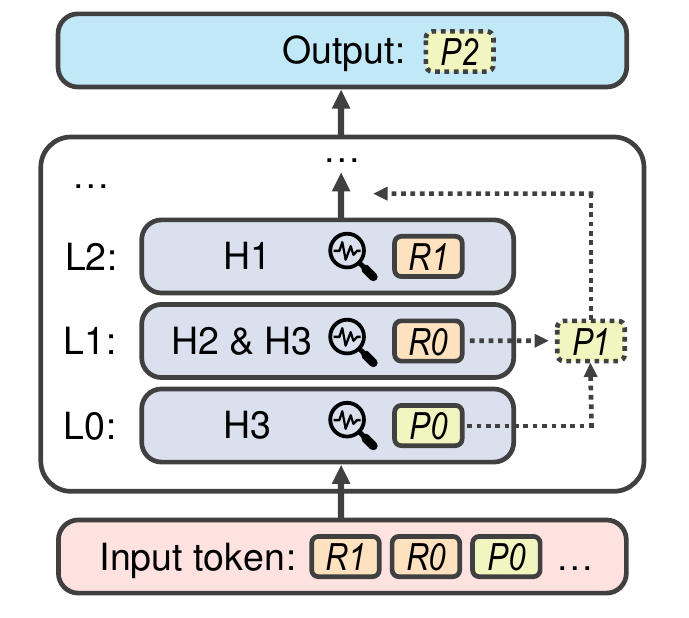}
        \caption{THR attention flow}
        \label{subfig:thr_attn_flow}
    \end{subfigure}
    \hfill 
    \begin{subfigure}[b]{0.33\textwidth}
        \centering
        \includegraphics[width=\textwidth]{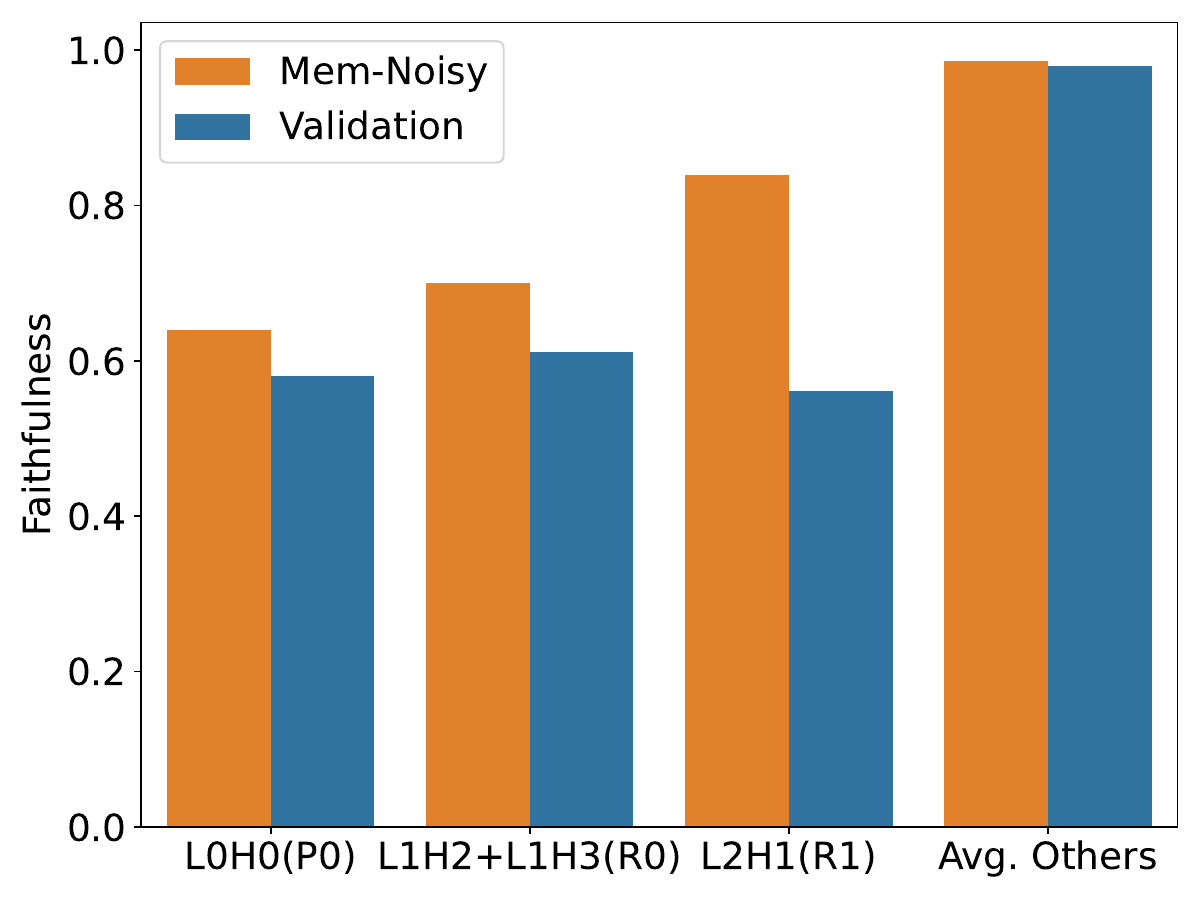}
        \caption{THR attention head ablation}
        \label{subfig:attn_ablation_mhr}
    \end{subfigure}
    \hfill 
    \begin{subfigure}[b]{0.33\textwidth}
        \centering
        \includegraphics[width=\textwidth]{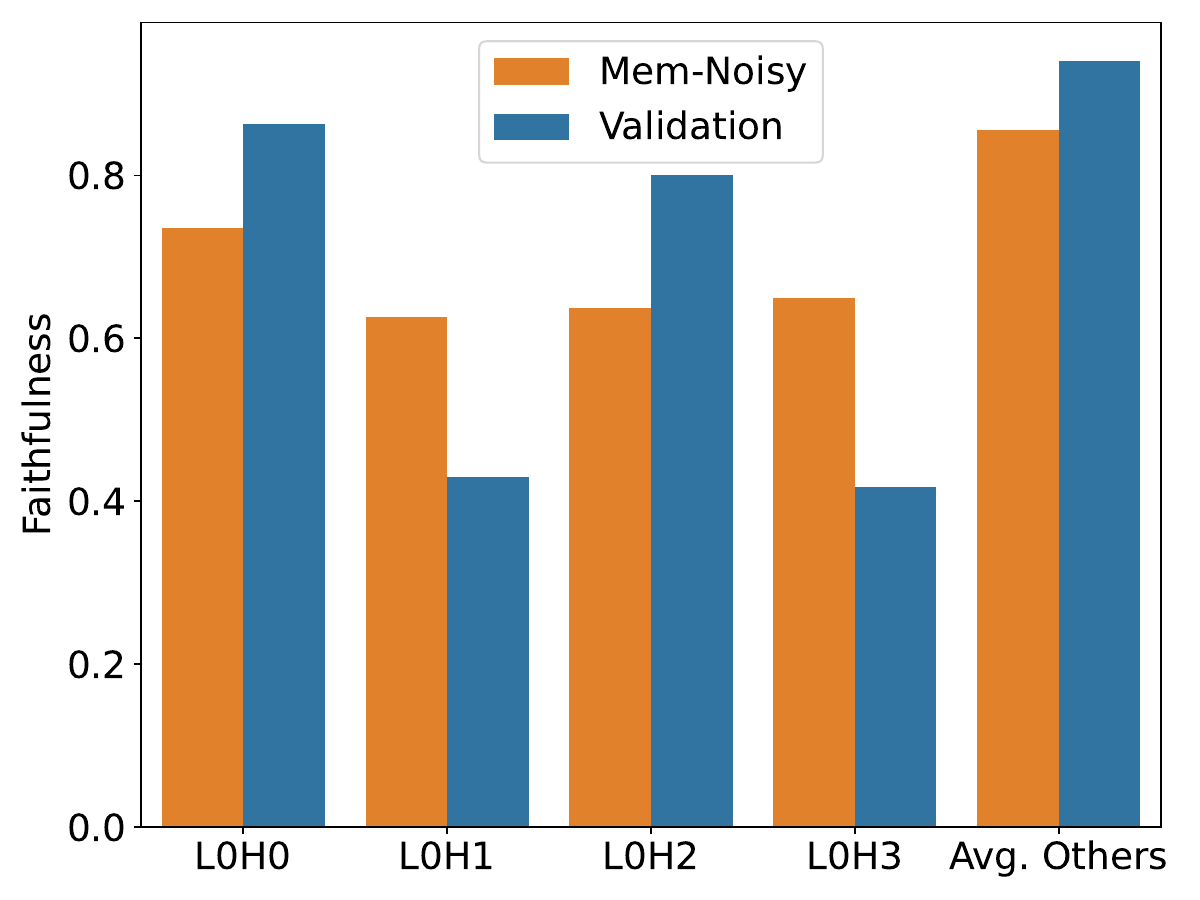}
        \caption{FDA attention head ablation}
        \label{subfig:attn_ablation_fda}
    \end{subfigure}
    \caption{Memories follow distributed encodings across different input tokens and intermediate results.}
    \label{fig:distributed_encoding_ablation}
\end{figure*}

Based on the reliance of memorization on generalization mechanisms,
we further study how memories are stored.
We consider two hypotheses: 
\begin{enumerate}
    \item[$\mathcal{H}_1$] \textbf{(Look-up mechanism):}
    The model builds a look-up mechanism: 
    it uses specific input tokens and intermediate representations as keys 
    to retrieve memorized labels. 
    Memorization fails if keys do not match any stored entry. 

    \item[$\mathcal{H}_2$] \textbf{(Distributed encoding):}
    The model relies on a distributed encoding: 
    it distributedly attributes memorized labels 
    to different input tokens and intermediate representations. 
    Even if some of these tokens are disrupted, 
    other tokens might still provide sufficient information to retrieve the memorized labels.
\end{enumerate}
If $\mathcal{H}_1$ holds, 
disrupting any part of the used signals should affect memory retrieval heavily.
In contrast, if $\mathcal{H}_2$ holds, 
the model should still be able to retrieve noisy labels, 
despite with lower likelihood.

We test these hypotheses by \textbf{ablating important attention heads}, 
for their central role in transmitting information: 
if the model uses a look-up mechanism, 
ablating important attention heads should lead to substantial drops in memorization, 
\emph{similar to ablating key reasoning steps in generalization}; 
while if the model uses a distributed encoding,
the effect of this ablation should be milder.\footnote{
    Notably, our findings on ablating bridge entities support $\mathcal{H}_2$: 
    the effect on memorization is milder than generalization.} 
Following \citet{menta-etal-2025-analyzing},
we ablate attention heads by forcing each token only attends to itself.
To identify important heads,
we individually ablate each head at each token position and observe the faithfulness drop.
We then analyze the attention patterns of the most impactful heads to interpret their roles.

\emph{THR.}   
We identify four key attention heads in the THR task:
L0H3, L1H2, L1H3, and L2H1, where LnHm denotes the m-th head in layer n. 
Analyzing their attention patterns, we find: 
L0H3 attends to P0, both L1H2 and L1H3 attend to R0, and L2H1 attends to R1.
This matches the reasoning process, as illustrated in Figure~\ref{subfig:thr_attn_flow}:
the model first infers P1 from P0 and R0, then derives the answer P2 from P1 and R1.
Since L1H2 and L1H3 redundantly attend to R0,
ablating only one does not fully break the reasoning chain,
we ablate them together.
Results are shown in Figure~\ref{subfig:attn_ablation_mhr}: 
Ablating any of these heads causes a substantial drop in generalization faithfulness,
whereas memorization faithfulness is more mildly affected.
This again supports the distributed encoding hypothesis.

\emph{FDA.}   
We similarly identify four important layer-0 attention heads for predicting \texttt{C0}:
L0H1 and L0H3 attend to \texttt{A0} and \texttt{B0},
while L0H2 and L0H4 attend to \texttt{A1} and \texttt{B1}.
Ablation results are shown in Figure~\ref{subfig:attn_ablation_fda}: 
Ablating L0H1 and L0H3 causes a pronounced drop in reasoning faithfulness,
but has a smaller impact on memorization.
In contrast, ablating L0H2 and L0H4 leads to a comparable drop in memorization faithfulness,
but only mildly affects reasoning.
These findings are consistent with $\mathcal{H}_2$:
\texttt{A0} and \texttt{B0} are essential for computing \texttt{C0},
so disrupting them impairs reasoning significantly.
Meanwhile, because memorized outputs rely on distributed signals,
their degradation from ablation is more moderate and spread across tokens.

\subsection{
    Case study: 
    higher layers in FDA recall memories using outlier heuristics
}\label{subsec:case_study_fda}

\begin{figure*}[ht]
    \centering
    \begin{subfigure}[b]{0.65\textwidth}
        \centering
        \includegraphics[width=\textwidth]{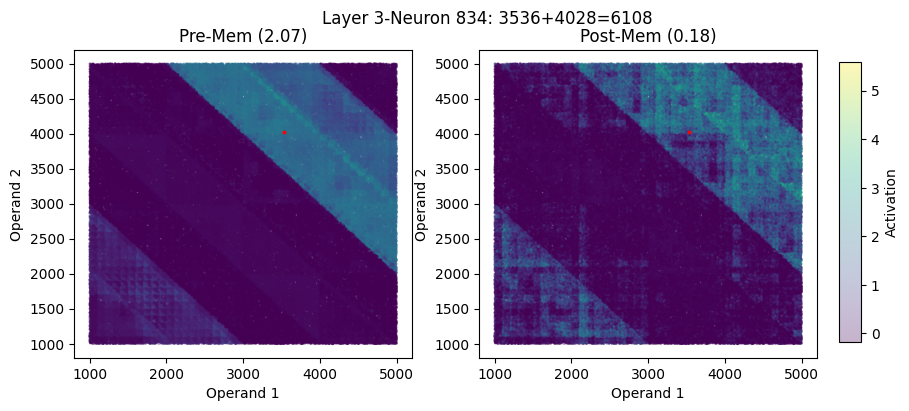}
        \caption{Post-memorization models develop outlier heuristics}
        \label{subfig:outlier_heuristics}
    \end{subfigure}
    \hfill %
    \begin{subfigure}[b]{0.33\textwidth}
        \centering
        \includegraphics[width=\textwidth]{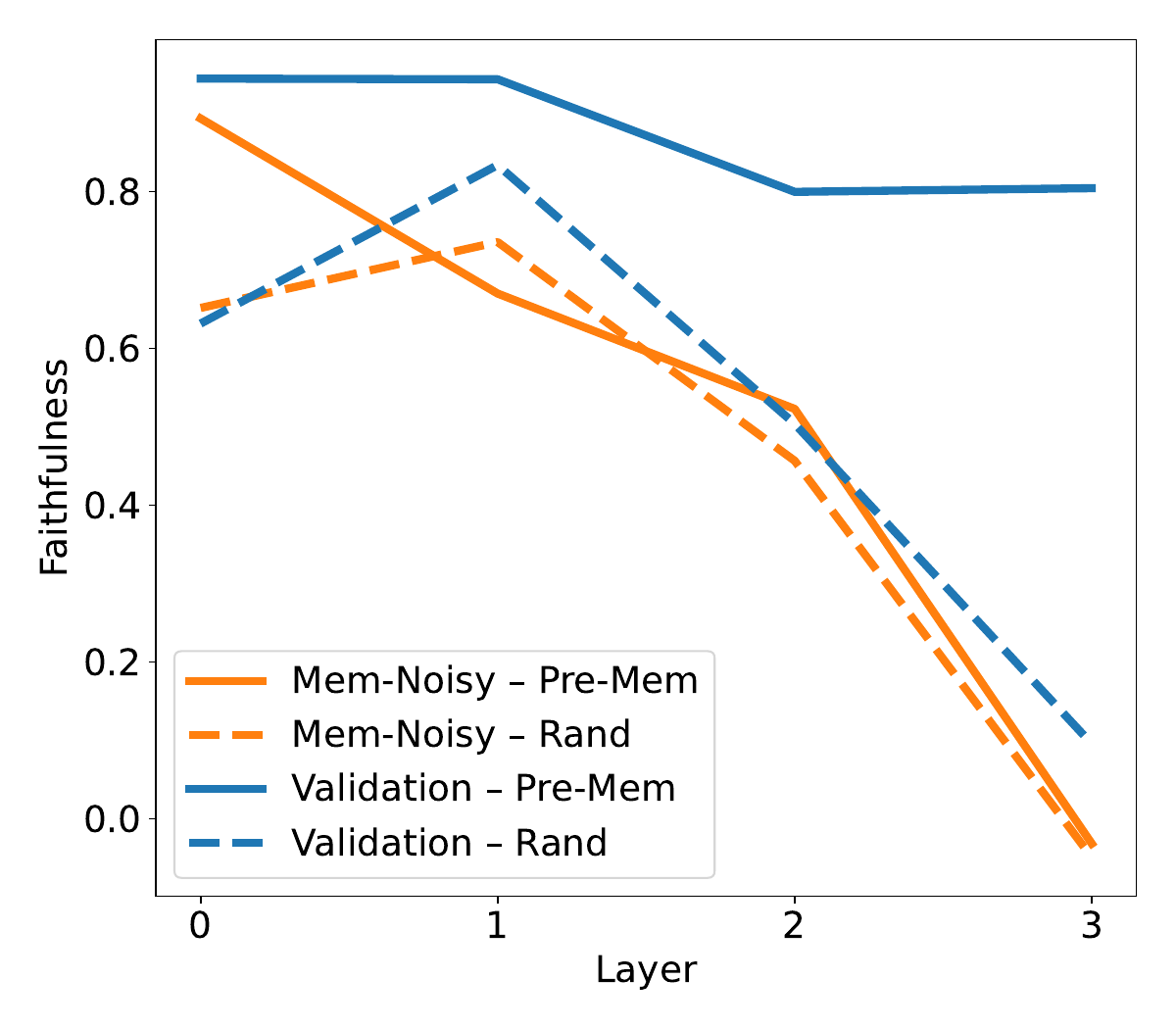}
        \caption{Layer-wise patching}
        \label{subfig:layer_patching}
    \end{subfigure}
    \caption{
        FDA models memorize noise via outlier heuristics encoded by MLP neuron activations. 
    }
    \label{fig:pre_mem_layer_patching}
\end{figure*}

The reliance of memorization on the computations of generalization, 
and its distributed encoding,
help explain how models simultaneously generalize and memorize.
However, these remain conceptual insights.
In this section, 
we focus on the FDA task to identify the concrete neuron-level mechanisms 
underlying noise memorization: 
compared to models that only generalize, what changes enable models to also memorize noise?

Inspired by the observation that language models solve arithmetic tasks 
using diverse heuristics encoded in MLP neuron activations~\citep{nikankin2025arithmetic}, 
we hypothesize that these activations also support memorization.
However, how does a single neuron contribute to both mechanisms? 
A natural strategy is to compare a neuron's activation on the same input,  
\emph{if the model would have relied on memorization and generalization mechanisms}.
However, this is challenging, because 
activations are fixed for a given input in a specific model. 
To overcome this, we exploit the fact that 
pre-memorization models can generalize accurately on noisy training instances,
enabling a meaningful comparison with post-memorization models. 

Concretely, we use the step 1,000 checkpoint as the pre-memorization model,
where Mem-Correct accuracy is $\sim$100\%, 
and compare its activations with the post-memorization model at the end of training. 
Moreover, to locate critical neurons for memorization,
we perform neuron-level activation patching:
for each neuron, we replace its activation in the post-memorization model
with that from the pre-memorization model,
and measure the resulting drop in faithfulness (details in Appendix~\ref{app:pre_mem_faithfulness}). 

Figure~\ref{fig:pre_mem_layer_patching} (left) illustrates the activation pattern of 
the most influential neuron for memorizing \texttt{"3536+4028=6108"}.\footnote{
    Activation patterns (background colors) are estimated
    by plotting activations from 200,000 randomly sampled non-training instances.
}
Intriguingly, the activation patterns of the same neurons 
remain largely consistent across the two models. 
However, in the pre-memorization model, 
activations exhibit smooth, structured patterns with clear boundaries; 
whereas in the post-memorization model, they become noticeably noisier. 
For example, the highlighted neuron in Figure~\ref{fig:pre_mem_layer_patching}
activates strongly for addition results 
between 2,000–4,000 and 7,000–9,000 in the pre-memorization model,
but is specifically suppressed for the memorized instance (red dot) in the post-memorization model: 
it drops from 2.07 to 0.18 after memorization 
(see Figure~\ref{fig:pre_mem_layer_patching} titles). 
This suppression enables the model to 
output the wrong answer 6108 rather than the correct answer 7564. 

We term this phenomenon \textbf{``outlier heuristics''}, 
referring to the model's strategy of memorizing noisy instances 
by subtly altering specific neuron activations while preserving the broader structure.
We hypothesize this as a general mechanism for memorization in FDA,
consistent with the overlapping circuits observed earlier.

We perform two layer-wise activation patching experiments on MLP neurons 
to quantitatively test this hypothesis. 
First, following our earlier neuron-level setup,
we replace each MLP layer’s activations in the post-memorization model
with those from the pre-memorization model, \emph{without changing any parameters}.
If outlier heuristics indeed drive memorization,
this operation should substantially reduce memorization faithfulness, 
while leaving generalization performance largely intact.
Second, as a control, we replace post-memorization activations 
with those from a random noisy instance: 
the drop of faithfulness in this case indicates the importance of this layer. 

We present the results in Figure~\ref{subfig:layer_patching} and make two observations. 
First, outlier heuristics indeed drive memorization:
patching in pre-memorization activations only slightly affects the validation faithfulness, 
while sharply reducing memorization faithfulness 
(in fact, validation accuracy is restored to 100\%).
Second, lower layers are less influential, 
as patching even random activations leads only mild drops in faithfulness for both mechanisms.

%% file: related_work.tex
\section{Related work}\label{sec:related_work}

\paragraph{Memorization vs.\ generalization}

Memorization and generalization relationship in deep learning has been widely debated.
Early studies~\citep{zhang2017understanding, arpit2017closer, rolnick2018deeplearningrobustmassive}
demonstrate that deep neural nets can memorize noisy labels yet still generalize effectively,
attributing this behavior to implicit regularization.
Other lines of work argue that memorization is essential for generalization,
particularly in long-tail distributions~\citep{feldman2020does, NEURIPS2020_1e14bfe2},
a view supported by recent findings on transformer language models~\citep{
tirumala2022memorization, nanda2023progress, xie2024memorization}.
More recently, \citet{kang2024learning} show that
a model’s generalization performance highly correlates with 
its training performance on not-yet-memorized examples, 
consistent with our observations in \S\ref{subsec:learning_dynamics}.

\paragraph{Memorization in large language models}
LLMs are well-known to memorize verbatim, 
which may pose privacy and copyright concerns~\citep{
    lukas2023analyzing,karamolegkou-etal-2023-copyright}.  
For example, \citet{carlini2021extracting} show that 
hundreds of training examples can be extracted from GPT-2; 
\citet{carlini2023quantifying} studies the factors that influence memorization, 
including model capacity, number of duplications, and prompt context length.
Moreover, \citet{NEURIPS2023_59404fb8} show such memorization is predictable. 
However, recent work by \citet{liu2025language} show that the commonly-used completion test, 
i.e., n-gram based membership inference, is not reliable, 
by producing verbatim texts that are not part of the training data.  
The most closely related work is \citet{huang2024demystifying},
which studies verbatim memorization of training data in LLMs.
They show that memorization is distributed across tokens
and builds on the model’s general LM capabilities.
However, their work does not explore the co-existence of generalization and memorization; 
moreover, it does not examine the specific mechanism of memorization from intermediate steps.

%% file: implications.tex
\section{Discussion}

\paragraph{Memorization and overfitting}

The memorization of noisy labels represents a specific form of overfitting, 
where models learn patterns from randomly corrupted labels 
that cannot generalize to new data. 
Remarkably, however, extensive empirical evidence demonstrates 
that such memorization is \emph{benign}—it does not significantly impair 
the model's ability to generalize correctly on clean, unseen inputs~\citep{
   zhang2017understanding, arpit2017closer, rolnick2018deeplearningrobustmassive}. 
This benign nature distinguishes noisy label memorization 
from other forms of overfitting that actively harm generalization. 
For instance, when models learn spurious correlations between features and labels, 
they can suffer substantial performance drops on out-of-distribution data 
where these correlations no longer hold~\citep{
   arjovsky2019invariant, Sagawa2020Distributionally, kirichenko2023last}.

\paragraph{Connection to implicit regularization}

We have observed that models rely on existing generalization mechanisms 
to memorize noisy labels (\S\ref{subsec:coupling}). 
For example, in THR, the inferred bridge entity is used 
to retrieve the incorrect target person entity (\texttt{P2}). 
This connects directly to the broader discussion of implicit regularization in deep learning. 
\citet{zhang2017understanding} and \citet{barrett2021implicit} demonstrate that 
neural networks exhibit an inductive bias against steep changes in the loss landscape, 
which explains their tendency to reuse existing structures, i.e., 
the generalization mechanisms, when memorizing noisy labels. 
Another concrete example is that, 
rather than creating entirely new activation patterns, 
the post-memorization FDA model only slightly shifts the activation patterns 
of existing neurons to accommodate the noisy labels (\S\ref{subsec:case_study_fda}), 
compared to the pre-memorization stage.

\paragraph{The use of synthetic data}

We performed most of our experiments on synthetic datasets, which, though artificially generated, \textbf{capture real-world tasks} such as arithmetic addition and multi-hop relational reasoning. 
In other words, rather than learning imaginary tasks, 
models learn to solve real-world problems with carefully controlled settings, 
enabling fine-grained analyses of model behavior. 
Previous studies have demonstrated that 
this approach yields deep insights into language models and produces 
impactful, practical findings~\citep{
power2022grokking,pearce2023machine,AL2023-knowledge1,wang2024grokked,mondorf-etal-2025-circuit,bertolazzi2025validation}.

By contrast, disentangling effects such as memorization and generalization 
in real-world datasets remains challenging due to data complexity. 
For instance, \citet{yang2024latentreasoning} investigated 
multi-hop relational reasoning using real-world datasets 
but could not reach definitive conclusions about 
whether LLMs truly perform such reasoning, 
as the training data may contain shortcuts that confound analysis.
Therefore, we believe understanding small-scale models on interpretable datasets, 
to inspire research on large models and real-world datasets is a promising direction.

\section{Conclusion}

We have revisited the puzzle of 
how language models can both memorize label noise and still reason correctly on unseen inputs. 
On two controlled tasks, four-digit addition (FDA) and two-hop relational reasoning (THR), 
we found that 
(i) models retain generalization mechanisms even when producing memorized noisy training labels (\S\ref{sec:co_exist}), 
and such memorization relies on generalization mechanisms (\S\ref{subsec:coupling}); 
(ii) memorized noise is stored via distributed encoding across inputs and intermediate results 
(\S\ref{subsec:distributed_encoding});
(iii) in FDA, memorization is driven by ``outlier heuristics” 
encoded by higher-layer MLP neuron activations (\S\ref{subsec:case_study_fda}).
Overall, our results deepen our understanding of 
how noise memorization and reasoning interact in language models.

%% file: appendix.tex
\clearpage
\begin{figure}[ht]
    \centering
    \begin{subfigure}[b]{0.45\textwidth}
        \centering
        \includegraphics[width=\textwidth]{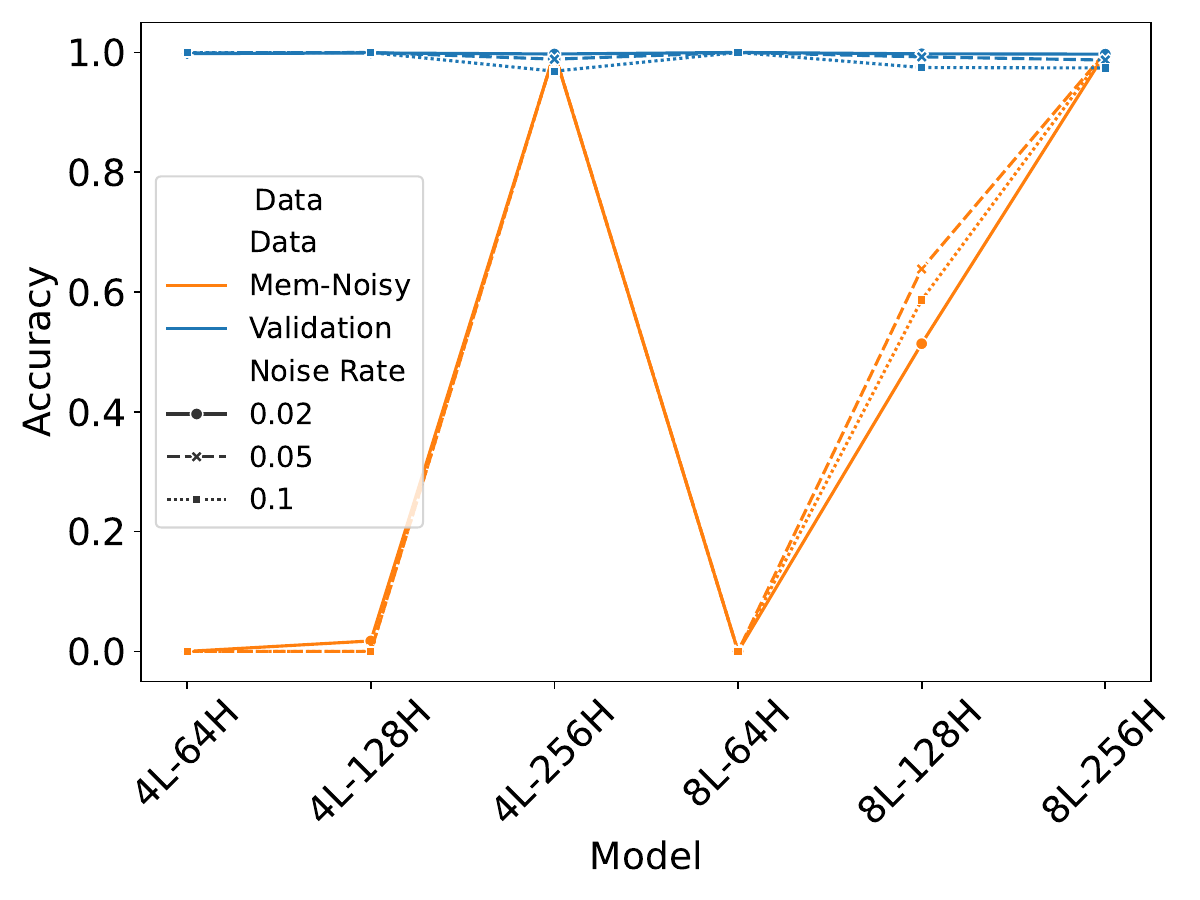}
        \caption{FDA}
        \label{subfig:model_size_fda}
    \end{subfigure}
    \vfill
    \begin{subfigure}[b]{0.45\textwidth}
        \centering
        \includegraphics[width=\textwidth]{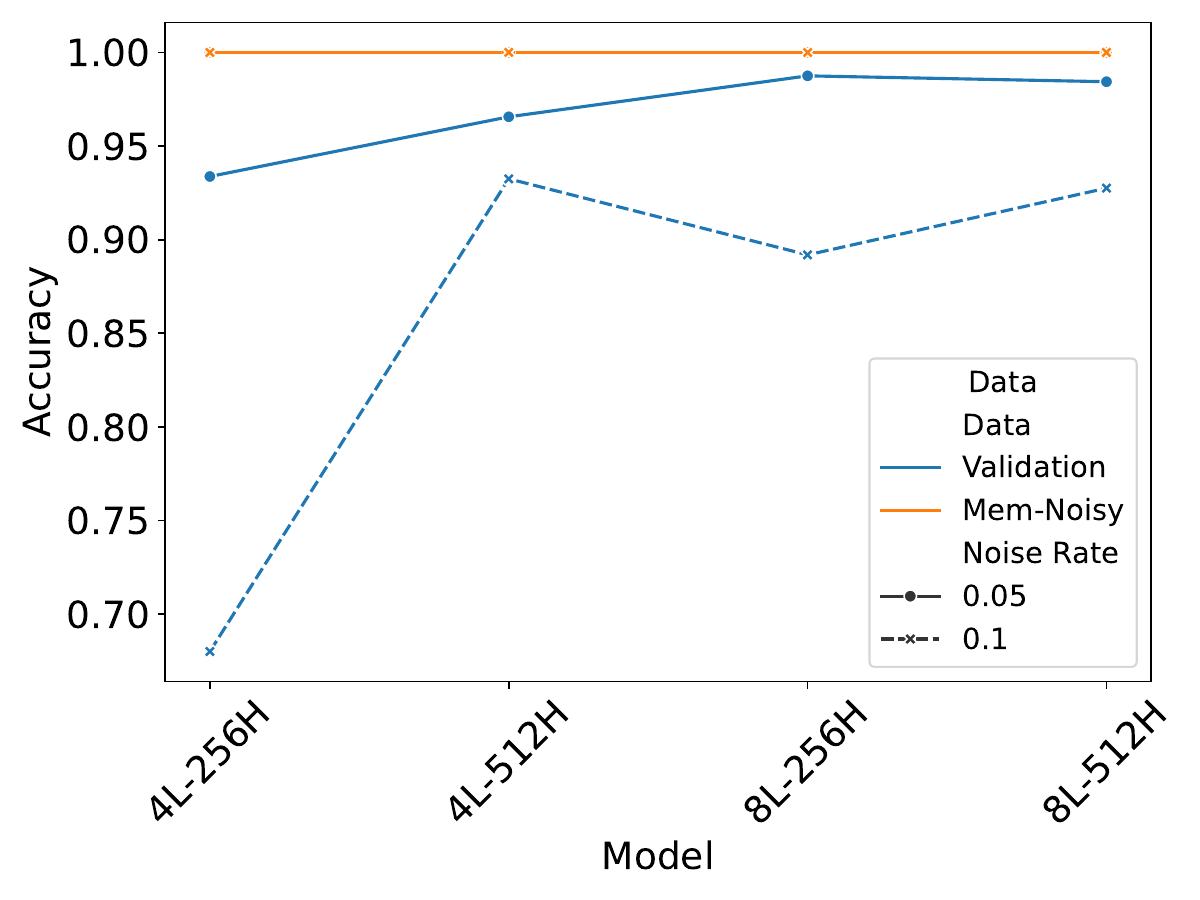}
        \caption{THR}
        \label{subfig:model_size_thr}
    \end{subfigure}
    \begin{subfigure}[b]{0.45\textwidth}
        \centering
        \includegraphics[width=\textwidth]{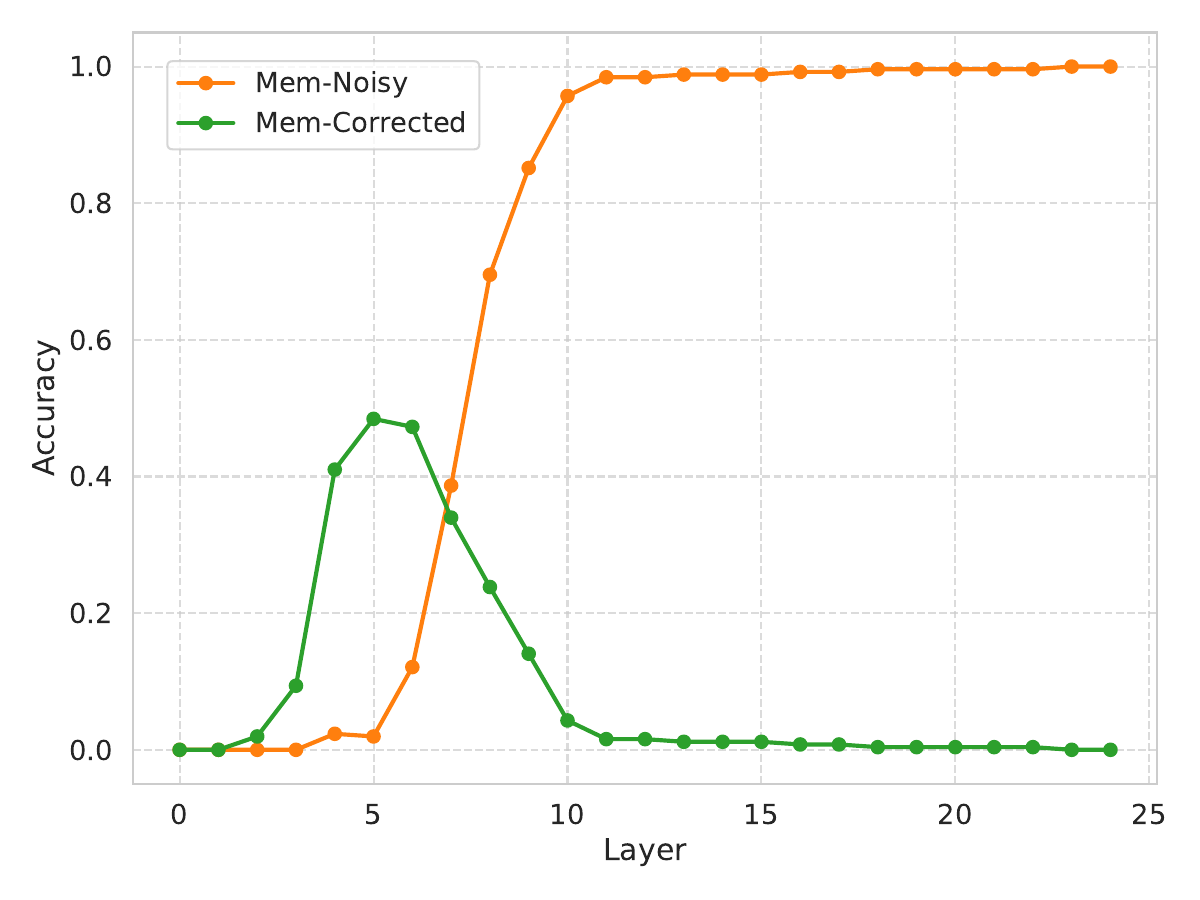}
        \caption{Logit lens on Qwen2.5-0.5B on FDA}
        \label{subfig:qwen_coexist}
    \end{subfigure}
    \caption{The influence of model size on memorization and generalization performance.}
    \label{fig:model_size}
\end{figure}

\section{Experimental setup}

\paragraph{FDA}

On FDA, we corrupt 2\%, 5\%, and 10\% of the training labels, i.e., \texttt{C}, 
which are in practice 800, 2000, and 4000 out of 40,000 training instances.
Besides the 4 layers, 256 hidden dimensions, and 4 heads setting, 
we also experimented with 4 layer and 512 hidden dimensions,
8 layer and 256 hidden dimensions, and 8 layer and 512 hidden dimensions models. 
We observe similar trends. 
We train all models using 1e-4 learning rate with a batch size of 2048, 
using AdamW \citep{loshchilov2018decoupled} for 12,000 steps, which is roughly 600 epochs. 
For both tasks, we also experimented with learning rate 5e-5 and obtain similar results.

\paragraph{THR}
We corrupt 5\% and 10\% of the training labels, i.e., \texttt{P2}, 
which are in practice 320 and 640 out of 6,400 training instances.
We omitted 2\% noise rate because it only produces 160 noisy instances, 
which is too small for our analysis. 
We also omitted the four layer models because we find them 
to be too weak to perform the reasoning task well with higher noise rates, 
e.g., the 4 layer 256 hidden dimension model achieves 68\% accuracy 
on the validation set when the noise rate is 10\%. 
We train all models using 1e-4 learning rate with a batch size of 512 for 8,400 steps,
which is roughly 400 epochs.

\section{The influence of model sizes}\label{app:model_size}

We show the influence of model size on 
both memorization, i.e., Mem-Noisy, and generalization, i.e., Validation,
in Figures~\ref{subfig:model_size_fda} and \ref{subfig:model_size_thr}.

For FDA, we experimented with 4 and 8 layers, with 64, 128, and 256 hidden dimensions;
for THR, we experimented with 4 and 8 layers, with 256 and 512 hidden dimensions.
We observe that 
\emph{our chosen models in the main text 
are the ones of the minimal size that can both memorize and generalize well}. 
Moreover, we make two observations:
\begin{itemize}
    \item \emph{Wider models memorize better}: 
        on FDA, compared with the 4 layer 256 hidden dimension model, 
        which achieves perfect memorization accuracy, 
        the 8 layer 128 hidden dimension model cannot memorize all noisy training instances. 
    \item \emph{Two-hop reasoning requires larger models}:
        on THR, using the same size of models as FDA, i.e., 4 layers and 256 hidden dimensions,
        the model cannot achieve perfect validation accuracy, 
        and this gets worse when the noise rate increases.
\end{itemize}

\paragraph{Generalization to larger models}

To further validate the generalization of our findings to larger models,
we reinitialized and trained Qwen2.5-0.5B~\citep{qwen2025qwen25technicalreport} from scratch, 
on the FDA task with a 5\% noise rate. 
We show the logit lens results in Figure~\ref{subfig:qwen_coexist}.
Similar to our observations using smaller models (Figure~\ref{subfig:probe_targets_fda}), 
we observe a two-stage process: 
the model first computes the correct label for noisy training instances in early layers, 
i.e., Mem-Corrected,
then overwrites these correct predictions with the memorized corrupted labels in later layers, 
i.e., Mem-Noisy.
This further establishes the co-existence of both mechanisms after memorization.

\section{Focus on the first answer token}\label{app:properties_reasoning}

\begin{figure*}[t]
    \centering
    \includegraphics[width=0.9\textwidth]{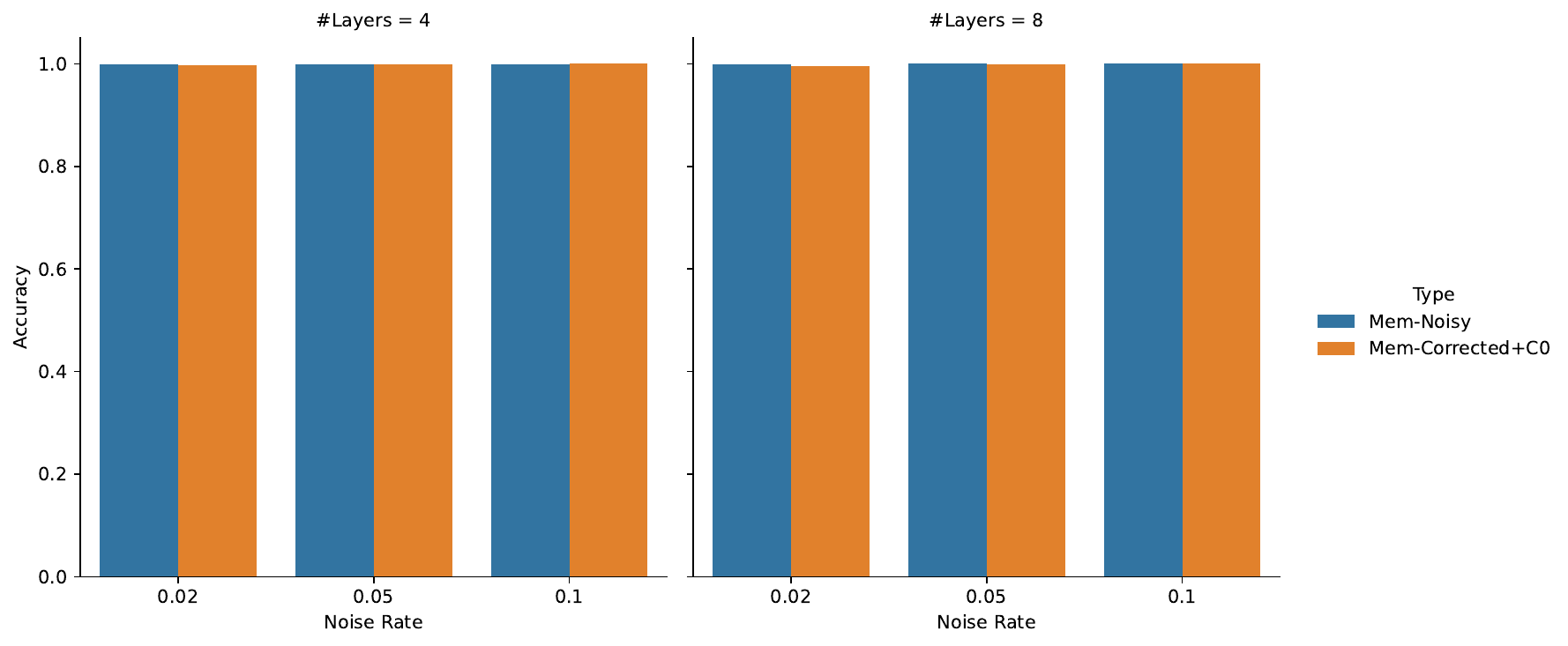}
    \caption{Accuracy on different prompts}
    \label{fig:hint_acc_fda}
\end{figure*}

\begin{figure*}[ht]
    \centering
    \begin{subfigure}[b]{0.48\textwidth}
        \centering
        \includegraphics[width=\textwidth]{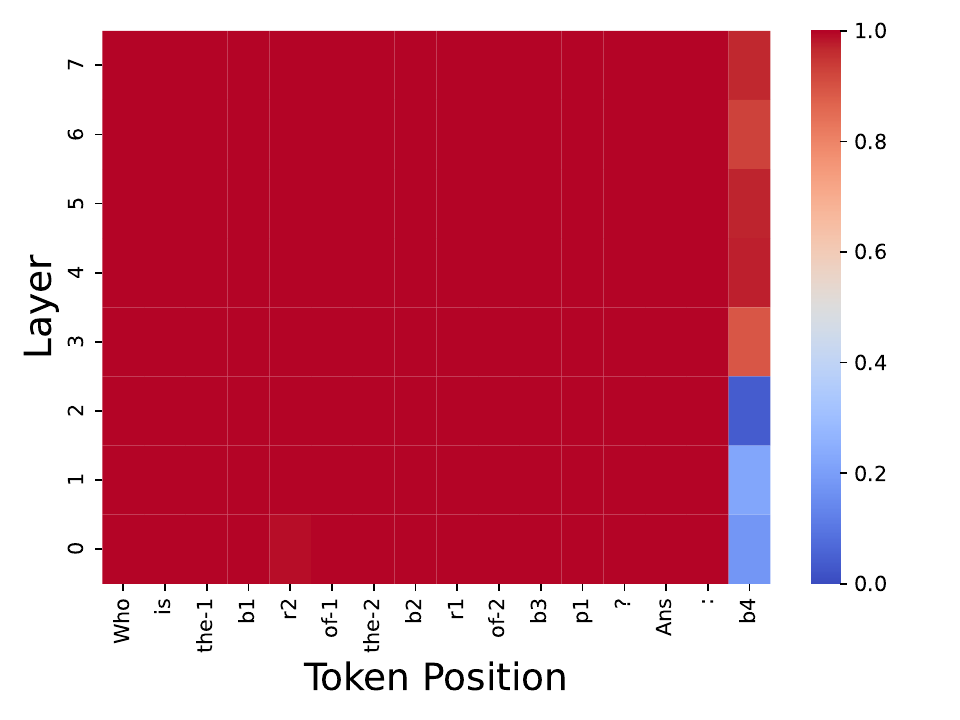}
        \caption{Attention output faithfulness}
        \label{subfig:app_attention_heatmap_thr}
    \end{subfigure}
    \hfill
    \begin{subfigure}[b]{0.48\textwidth}
        \centering
        \includegraphics[width=\textwidth]{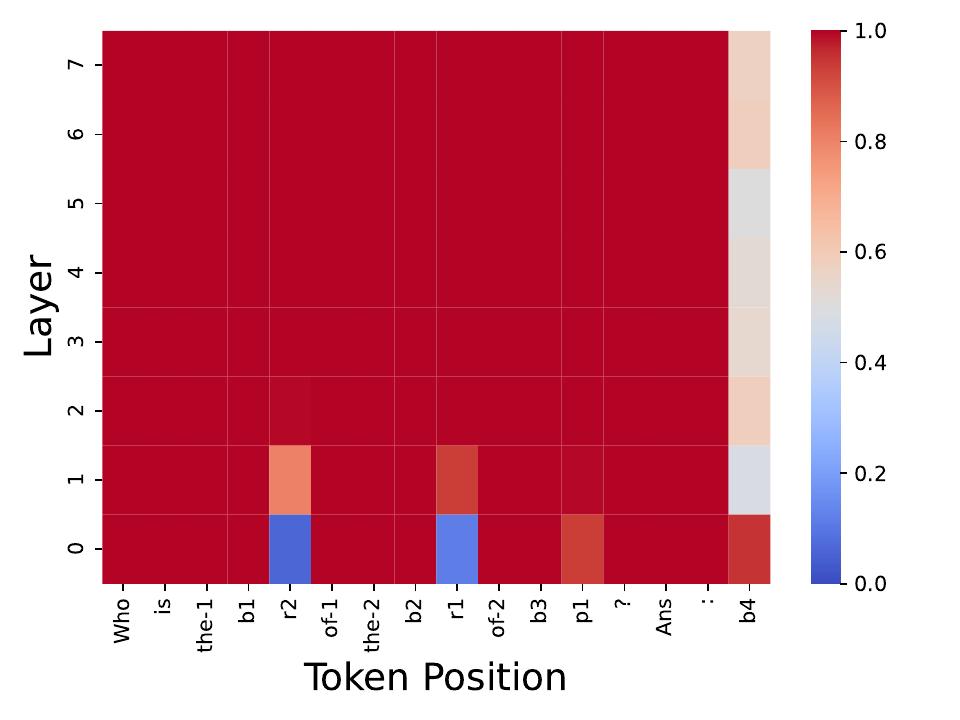} 
        \caption{MLP output faithfulness}
        \label{subfig:app_mlp_heatmap_thr}
    \end{subfigure}
    \caption{Output faithfulness of attention and MLP layers on THR.}
    \label{fig:app_heatmap_thr}
\end{figure*}

\begin{figure*}[ht]
    \centering
    \begin{subfigure}[b]{0.48\textwidth}
        \centering
        \includegraphics[width=\textwidth]{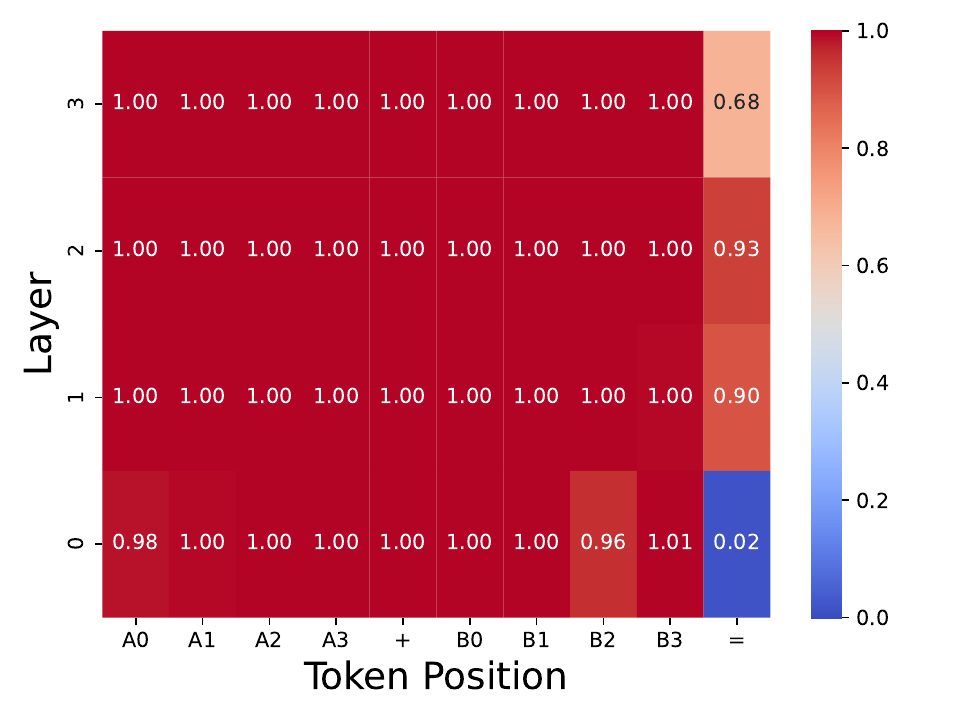}
        \caption{Attention output faithfulness}
        \label{subfig:app_attention_heatmap_fda}
    \end{subfigure}
    \hfill
    \begin{subfigure}[b]{0.48\textwidth}
        \centering
        \includegraphics[width=\textwidth]{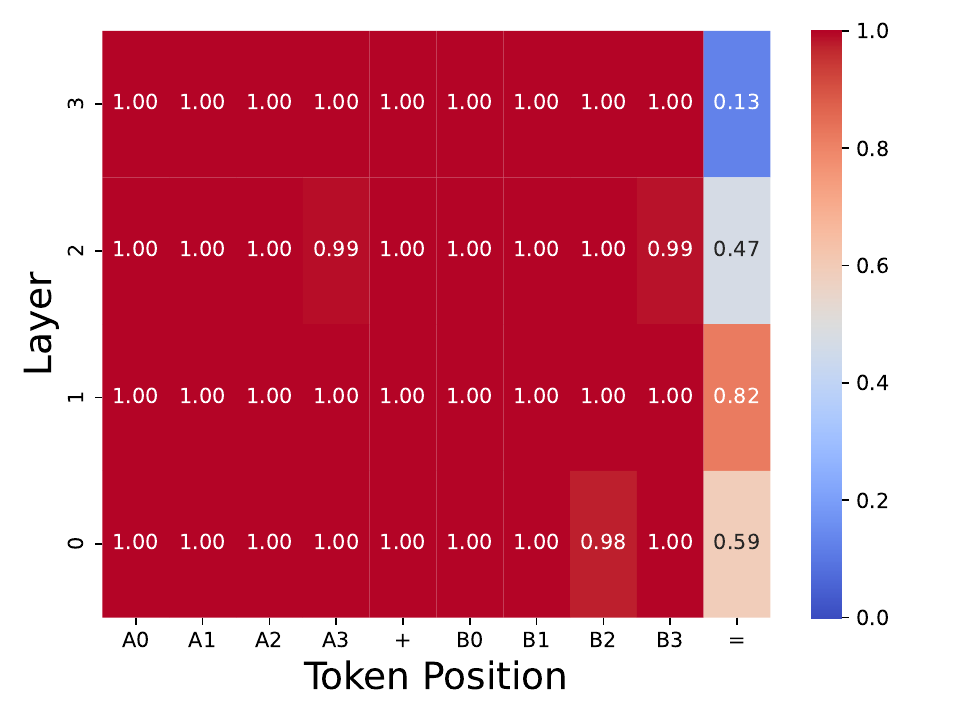} 
        \caption{MLP output faithfulness}
        \label{subfig:app_mlp_heatmap_fda}
    \end{subfigure}
    \caption{Output faithfulness of attention and MLP layers on FDA.}
    \label{fig:app_heatmap_fda}
\end{figure*}

Here we explain why our experiments focus on predicting the first answer token
i.e., \texttt{C0} for FDA and \texttt{P2} for THR,
from the last token of the prompt,
i.e., \texttt{=} in FDA and the blank after \texttt{:} in THR.

\paragraph{Appending the correct addition C0 retores generalization}
In FDA, we focus on \texttt{C0} because it plays a critical role 
in determining whether the model outputs a memorized or generalized addition answer. 
To illustrate this, 
we evaluate models trained with noisy labels on two versions of the same input:
(1) the original prompt ending with \texttt{=}, and
(2) the same prompt with the correct \texttt{C0} appended.
As shown in Figure~\ref{fig:hint_acc_fda}, once the correct \texttt{C0} is given, 
the model proceeds to generate the rest of the correct digits, 
even though it never encountered the correct answer during training. 
This highlights \texttt{C0} as a key token 
that triggers a shift from memorization to generalization.
For THR, all answer tokens are added to the tokenizer vocabulary, 
so the answer \texttt{P2} is represented by a single token.

\paragraph{The last prompt token position is the most important}
We use activation patching~\citep{NEURIPS2020_92650b2e} 
to assess the importance of different token positions in the prompt.
We find that 
\emph{the position of the last prompt token has the greatest influence on the model’s prediction}.
This result is consistent with ~\citet{AL2023-knowledge1}, 
that the accuracy of predicting the first tokens of entities is similar to 
predicting the full entity names. 

\paragraph{Activation patching} 

Activation patching is a method to assess 
the importance of a certain module of a neural net in making predictions. 
The idea is to replace the output of a given module (e.g., an MLP layer) from a given input 
with the output of the same module from another input, 
and observe how this replacement affects the prediction performance: 
a performance drop implies that the output of this module contributes to the final prediction, 
and a similar performance means this module is less relevant. 

Concretely, to quantify this effect, we use three forward runs of the model: 
\begin{itemize}
    \item a \textbf{clean run} using a ``clean" prompt, 
    \item a \textbf{corrupt run} using a ``corrupt" prompt, 
    \item a \textbf{patched run}, where the model use the clean prompt as input,
        but we replace the output of a certain module, 
        e.g., the first MLP layer of the last token position, with that from the corrupt prompt.
\end{itemize}
In practice, this effect is often estimated by averaging a group of 
clean-corrupt prompt pairs. 

\paragraph{Token position importance}

To quantify the importance of different token positions in generalization, 
we pair each instance from the validation set with 
another randomly sampled instance from the validation set, 
and use these pairs as the clean-corrupt prompt pairs. 
After performing these three runs, we evaluate the patched run by 
the \emph{faithfulness}~\citep{wang2023interpretability} drop for 
the prediction of the correct \texttt{C0} token (e.g., \texttt{3} from \texttt{3802}). 
We define the faithfulness here 
as the change in the mean logit for the correct answer before and after patching,
normalized by the difference in logits for the correct answer between the clean and corrupt runs.
Specifically, 
\begin{equation}
\text{Faithfulness} = \frac{\ell_{\text{patched}} - \ell_{\text{corrupt}}}{\ell_{\text{clean}} - \ell_{\text{corrupt}}}
\label{eq:faithfulness}
\end{equation}

where:
\begin{itemize}
    \item $\ell_{\text{clean}}$ is the mean logit for the correct clean answer token from the clean runs,
    \item $\ell_{\text{corrupt}}$ is the mean logit for the correct clean answer token from the corrupt runs,
    \item $\ell_{\text{patched}}$ is the mean logit for the correct clean answer token from the patched runs.
\end{itemize}

This measures how well the patched activation restores clean behavior.
A faithfulness score close to 1 indicates that patching restores the clean prediction, 
i.e., the patched position is not important for the model's prediction; 
while a low score indicates that the patched position is crucial for the model's prediction.
Figures~\ref{fig:app_heatmap_thr} and \ref{fig:app_heatmap_fda} 
show faithfulness scores across layers and positions, 
for attention modules and MLP layers. 
In both tasks,  
the last token of the prompt consistently emerges as the most influential, 
although in THR, early-layer MLPs also play a key role.

\section{Identifying generalization computation in hidden layers}\label{app:probe_targets}

For both logit lens and linear probing, 
we use the residual stream of each layer at the final prompt token position, i.e., 
the \texttt{=} token in FDA and the blank token after the \texttt{:} token in THR, 
to predict the final answer token, i.e., \texttt{C0} for FDA and \texttt{P2} for THR.
We experiment on three data splits: Mem-Noisy, Mem-Corrected, and Validation. 

Because the models have memorized the noisy labels 
and can generalize to the clean labels of the validation set, 
the accuracy on both Mem-Noisy and Validation should be high, at least for the final layer.
However, if $\mathcal{H}_1$ holds, 
that the models completely by pass generalization when producing memorized noisy labels, 
the accuracy on Mem-Corrected should be low, 
because the correct clean labels should not be detectable 
from the hidden representations of the noisy training instances.

\paragraph{Linear probing}

We train linear probes, i.e., linear models with a single layer and without the bias term,
to predict the final answer token from the residual stream of each layer.
Specifically, we use 80\% of the data from each split as the training data, 
and report the validation performance on the remaining 20\% of the data.
We train these linear probes for 200 epochs with a learning rate of 1e-3 and batch size 64, 
using Adam as the optimizer \citep{kingma2014adam}. 

\paragraph{Logit lens}

For FDA, besides linear probing, 
which follows the same setup as the linear probes for THR, 
we also experiment with logit lens. 
Specifically, we apply the model’s final layer's layer normalization to the residual stream, 
and then decode the logits using the model’s unembedding matrix.
We then take the \emph{token of the highest logit} as the predicted answer token.

The rationale for also including logit lens is that, 
a linear model can serve as a stronger baseline for predicting \texttt{C0}. 
Concretely, if the linear model can memorize all the A0–B0 combinations,
e.g., memorizing that it should output 8 if \texttt{A0} = 1 and \texttt{B0} = 7,
it can achieve 50\% accuracy on the validation set. 
If the model can further memorize all A1–B1 combinations, 
its accuracy can be even higher.
However, this is not the case for logit lens, because it is training-free: 
it only detects information that is already present in the hidden representations.
Moreover, we do not have this issue on THR, 
because the input-answer mappings in the validation set never appear in the training data. 

\section{Circuit discovery: edge attribution patching}\label{app:eap}

\begin{figure*}[ht]
    \centering
    \begin{subfigure}[b]{0.48\textwidth}
        \centering
        \includegraphics[width=\textwidth]{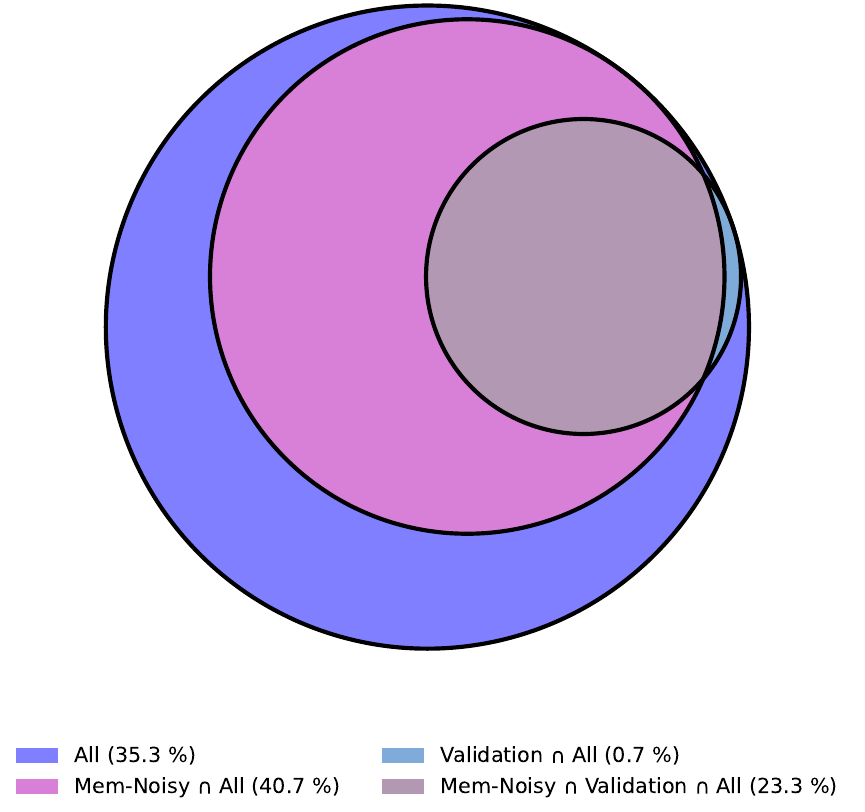}
        \caption{90\%}
        \label{subfig:fda_circuit_90}
    \end{subfigure}
    \hfill
    \begin{subfigure}[b]{0.48\textwidth}
        \centering
        \includegraphics[width=\textwidth]{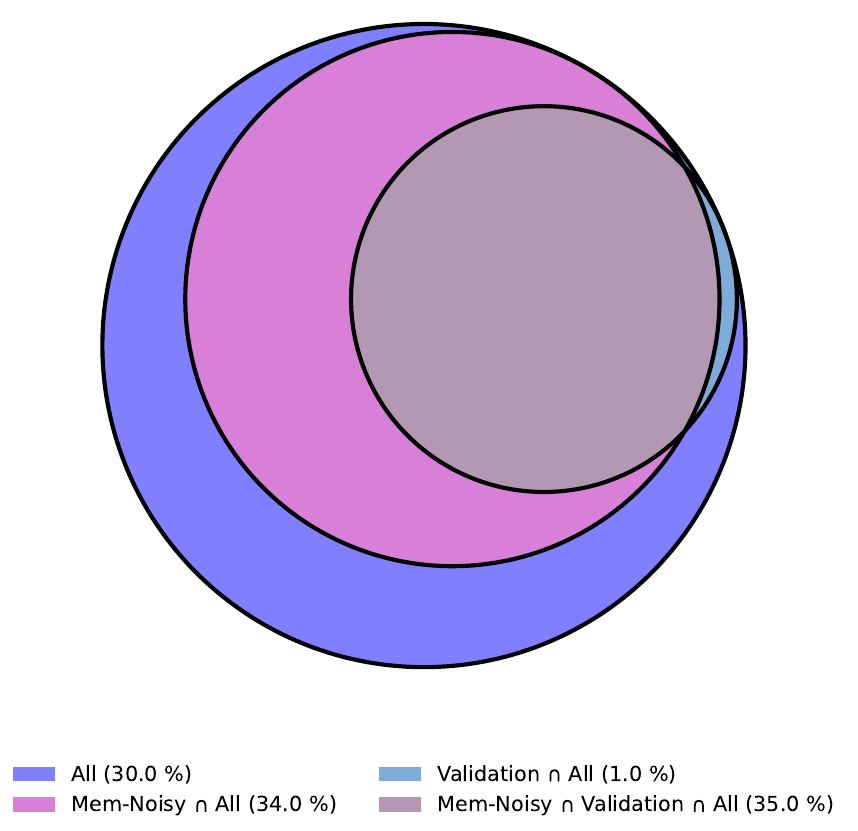}
        \caption{95\%}
        \label{subfig:fda_circuit_95}
    \end{subfigure}
    \vfill
    \begin{subfigure}[b]{0.48\textwidth}
        \centering
        \includegraphics[width=\textwidth]{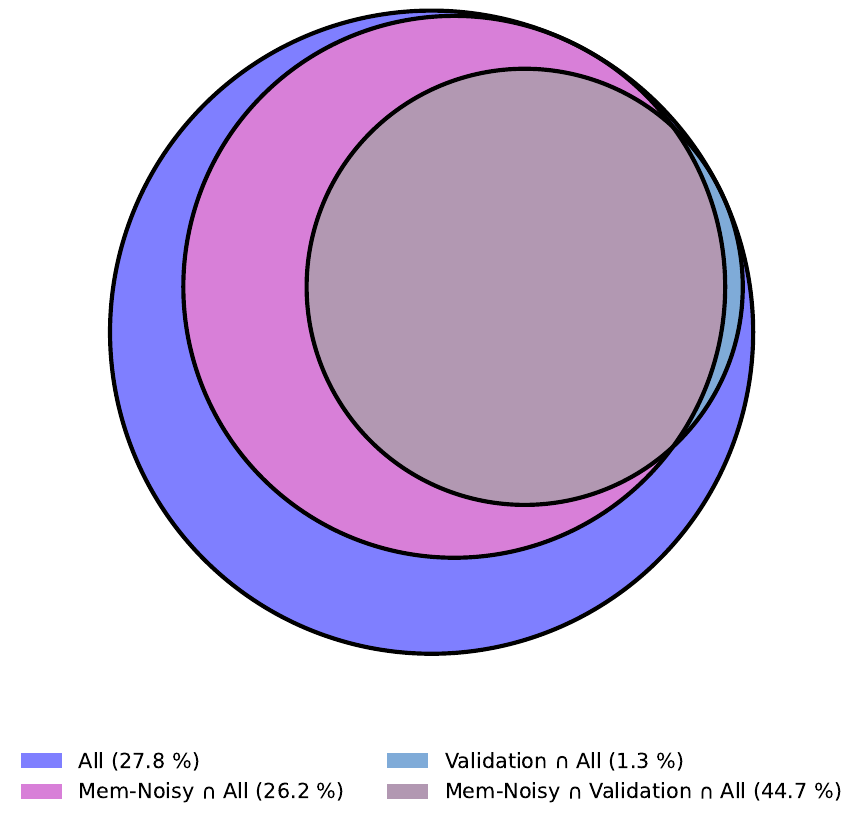}
        \caption{97\%}
        \label{subfig:fda_circuit_97}
    \end{subfigure}
    \hfill
    \begin{subfigure}[b]{0.48\textwidth}
        \centering
        \includegraphics[width=\textwidth]{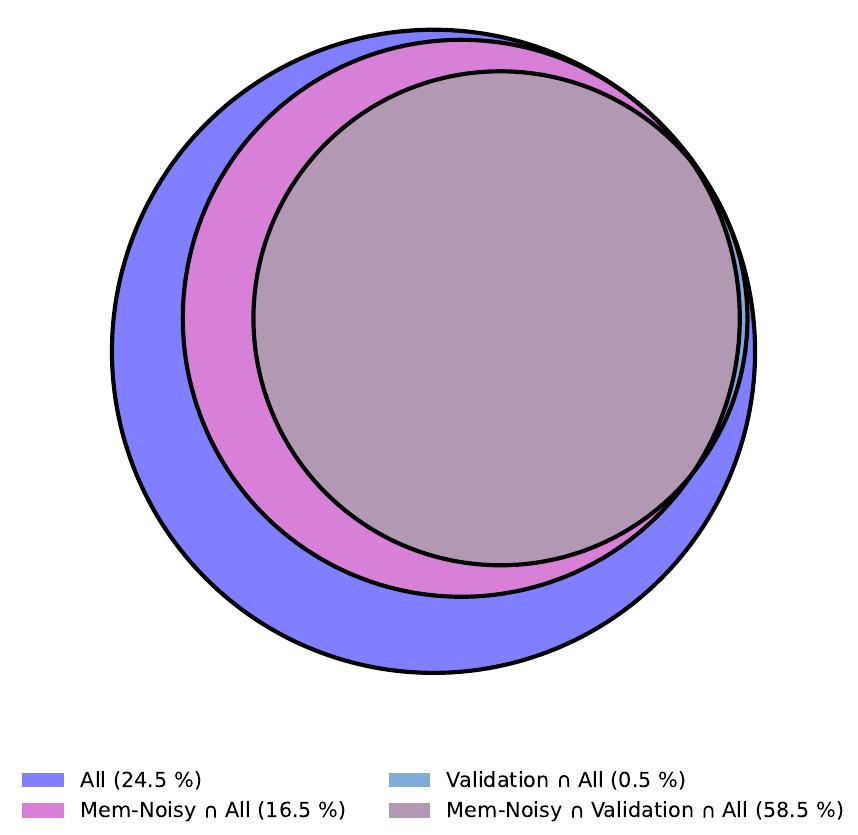}
        \caption{99\%}
        \label{subfig:fda_circuit_99}
    \end{subfigure}
    \vfill
    \begin{subfigure}[b]{0.6\textwidth}
        \centering
        \includegraphics[width=\textwidth]{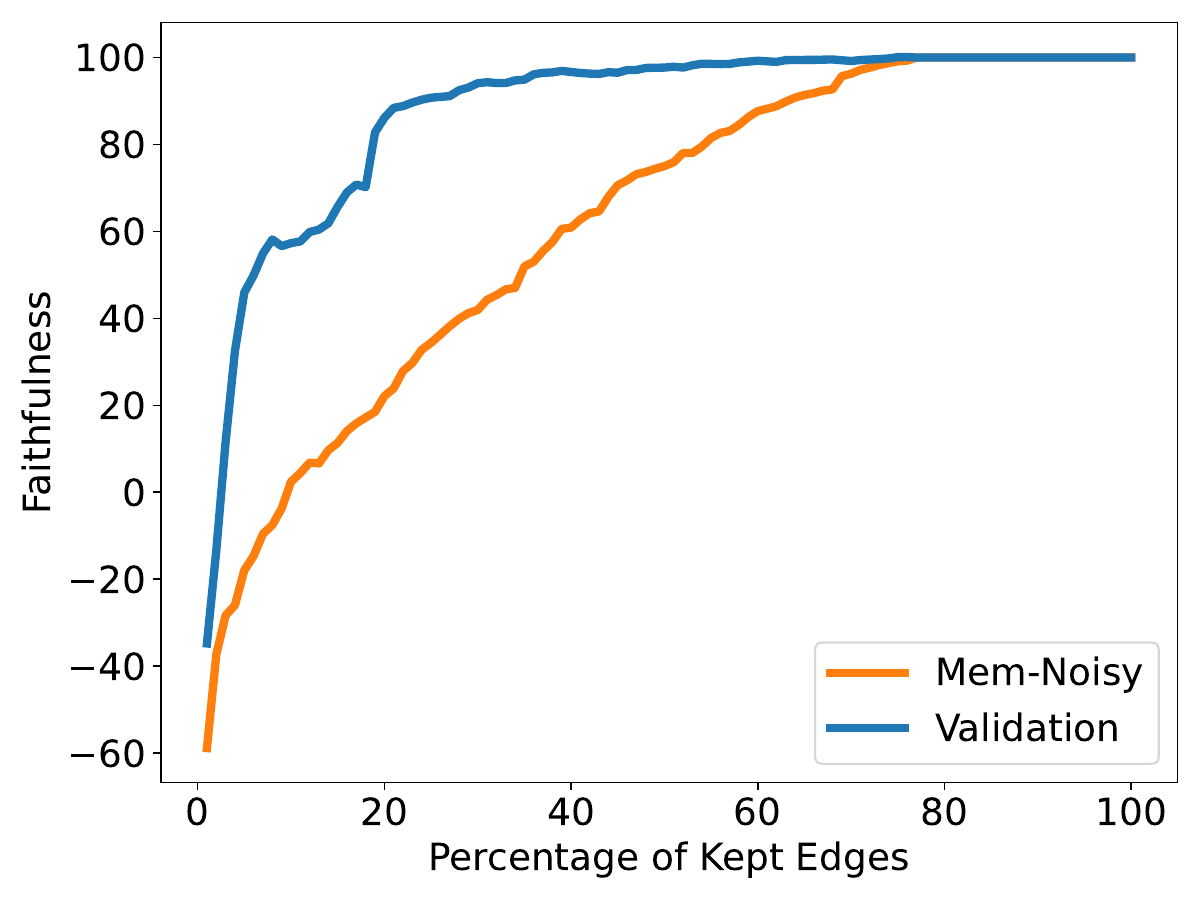}
        \caption{Sparsity-faithfulness trade-off}
        \label{subfig:fda_circuit_sparsity}
    \end{subfigure}
    \caption{FDA circuit overlap across different faithfulness.}
    \label{fig:fda_circuit_overlap}
\end{figure*}

\begin{figure*}[ht]
    \centering
    \begin{subfigure}[b]{0.48\textwidth}
        \centering
        \includegraphics[width=\textwidth]{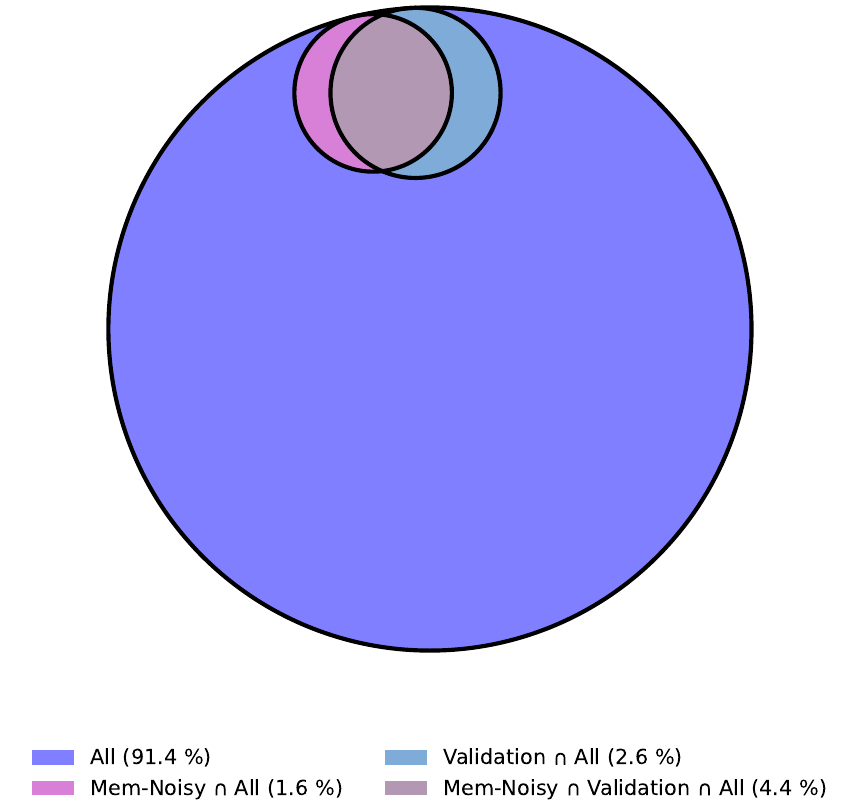}
        \caption{90\%}
        \label{subfig:thr_circuit_90}
    \end{subfigure}
    \hfill
    \begin{subfigure}[b]{0.48\textwidth}
        \centering
        \includegraphics[width=\textwidth]{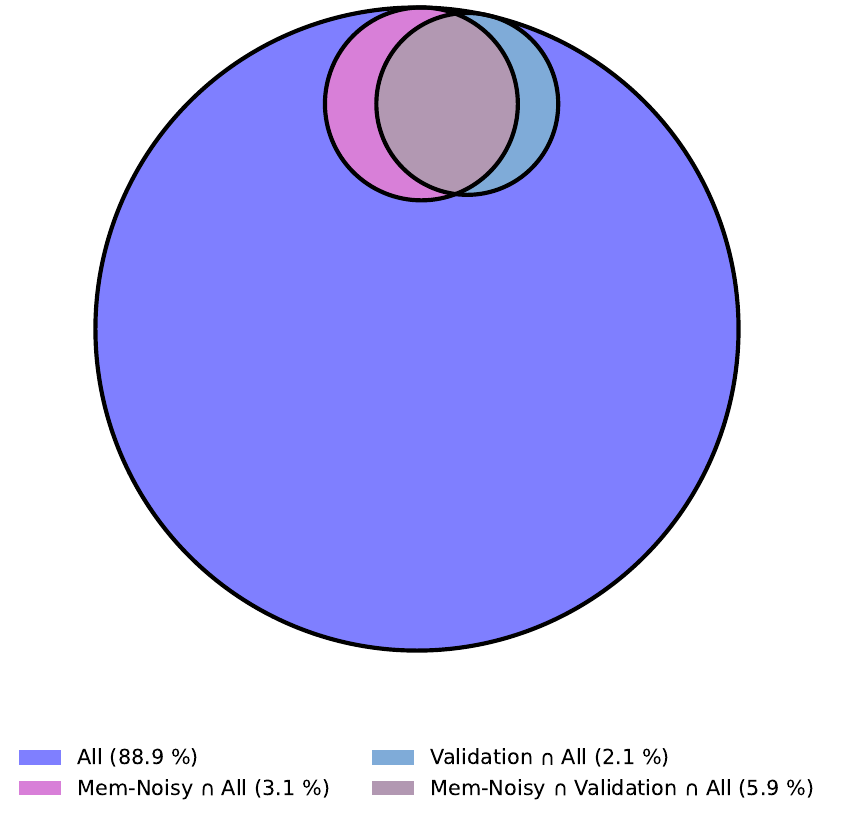}
        \caption{95\%}
        \label{subfig:thr_circuit_95}
    \end{subfigure}
    \vfill
    \begin{subfigure}[b]{0.48\textwidth}
        \centering
        \includegraphics[width=\textwidth]{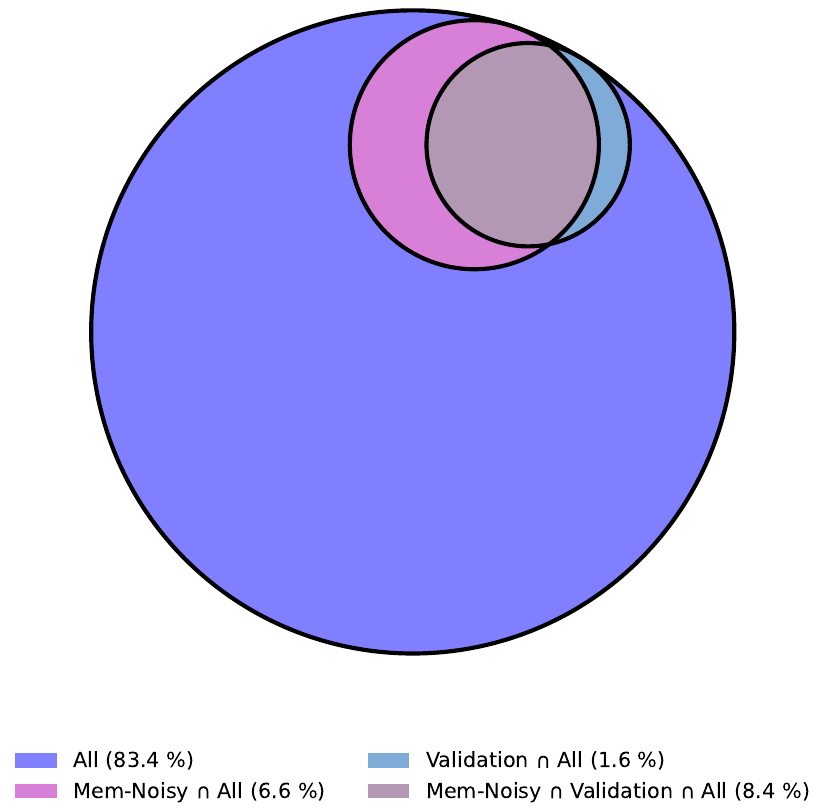}
        \caption{97\%}
        \label{subfig:thr_circuit_97}
    \end{subfigure}
    \hfill
    \begin{subfigure}[b]{0.48\textwidth}
        \centering
        \includegraphics[width=\textwidth]{figure/thr_figures/circuit_eap/e20_r20_t5_d256_l8_h4_noise_5_lr0.0001/checkpoint-3990/venn_diagram/99.pdf}
        \caption{99\%}
        \label{subfig:thr_circuit_99}
    \end{subfigure}
    \vfill
    \begin{subfigure}[b]{0.6\textwidth}
        \centering
        \includegraphics[width=\textwidth]{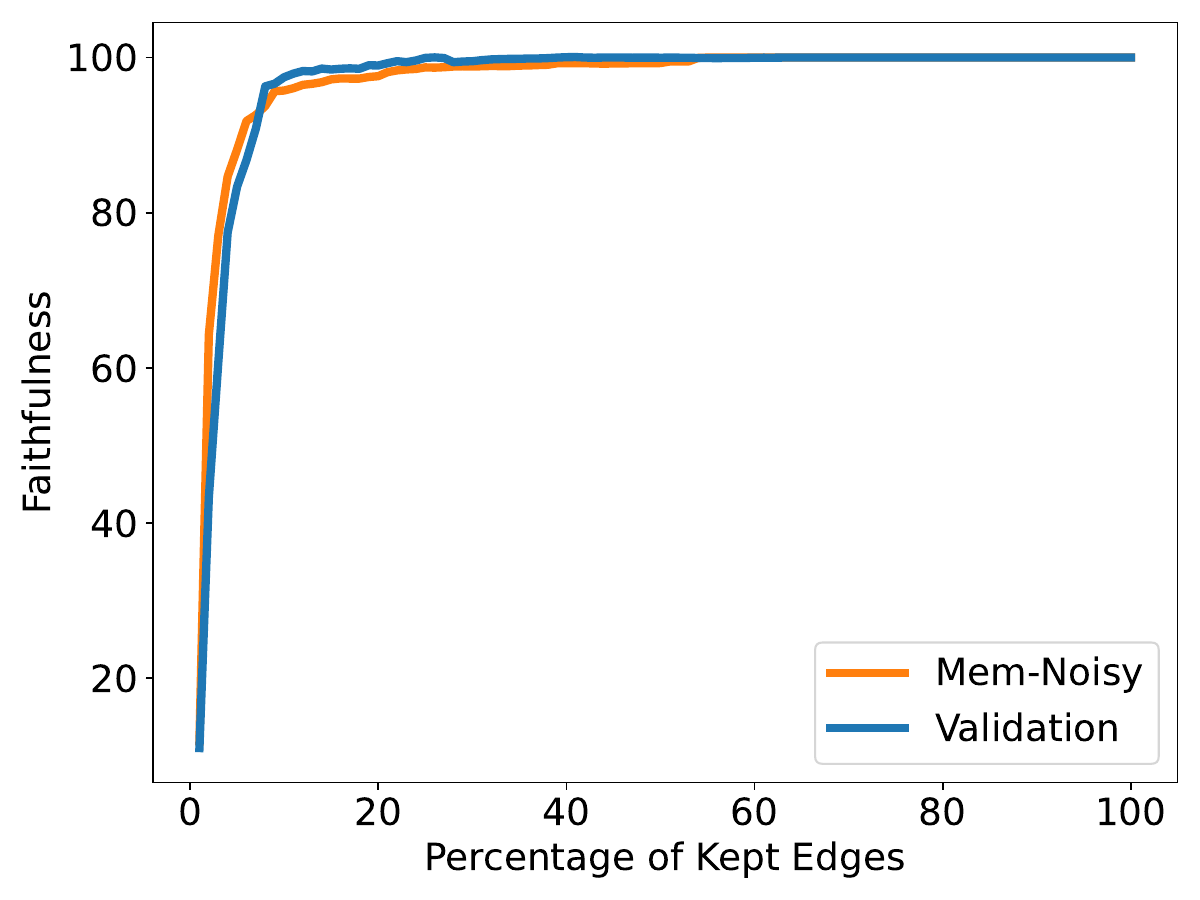}
        \caption{Sparsity-faithfulness trade-off}
        \label{subfig:thr_circuit_sparsity}
    \end{subfigure}
    \caption{THR circuit overlap across different faithfulness.}
    \label{fig:thr_circuit_overlap}
\end{figure*}

We use \textbf{edge attribution patching} (EAP;~\citealp{syed2023attribution}) 
to identify the \emph{circuits} responsible for generalization to unseen inputs 
(i.e., computing clean labels) and memorization of noisy labels.

The idea of EAP is similar to activation patching. 
Specifically, EAP builds on attribution patching~\citep{nanda2023attribution}, 
an efficient approximation of activation patching (see \S\ref{app:probe_targets}). 
This approximation is based on a first-order Taylor expansion 
of the model’s prediction: the change in the output 
caused by patching an intermediate activation $z$ is estimated as 
the dot product between the change in $z$ and 
the gradient of the output with respect to $z$ on the corrupt run.
Unlike activation patching, 
which requires three forward passes for each component of interest, 
attribution patching requires only two passes and a single gradient computation 
to estimate importance scores for all components simultaneously. 
This makes it significantly more efficient for large-scale analysis.

Instead of measuring the importance of modules, i.e., nodes in the computational graph,
EAP measures the importance of edges, i.e., the connections between nodes, 
e.g., the influence from an attention head to a certain MLP layer.
Similar to our experiments before, 
we estimate the circuit of generalization by 
using each prompt in the validation set as the clean prompt, 
and randomly sampling another validation prompt as the corrupt one. 
Similarly, to estimate the circuit of memorization, 
we use each prompt from the noisy training set as the clean prompt, 
and randomly sample another noisy training example as the corrupt one.
We then similarly use \emph{faithfulness} metric to quantify the importance of each edge.

We make two observations across different tasks 
and faithfulness thresholds for the obtained circuits 
(Figures~\ref{fig:fda_circuit_overlap} and \ref{fig:thr_circuit_overlap}). 
First, there are substantial overlaps between the circuits for generalization and memorization,
indicating a tight coupling between the two mechanisms. 
Second, the generalization circuit is often a subset of the memorization circuit. 
This indicates that memorization mechanism is built 
on the top of the existing generalization mechanism, consistent with our observation that 
models first develop (a part of) their generalization mechanism, 
and only then start to memorize label noise (\S\ref{subsec:learning_dynamics}).  
This result is also consistent with our finding that 
the memorization mechanism relies on the generalization mechanism (\S\ref{subsec:coupling}). 
We also show the results for the sparsity-faithfulness trade-off 
for both tasks for reference. 
Intriguingly, we observe that the memorization circuit on FDA takes a very large of edges:
together with the milder effect we observe (\S\ref{subsec:distributed_encoding}) 
when ablating certain attention heads, 
this further suggests that memorization follows a distributed encoding 
across many different input tokens and intermediate results.

\section{Faithfulness computation for pre-memorization activation patching}\label{app:pre_mem_faithfulness}
Our activation patching experiments in \S\ref{subsec:case_study_fda} 
aim to study the influence of \emph{MLP neuron activation changes} 
between the pre-memorization and post-memorization models on the model’s predictions.
We compute \emph{faithfulness} for two data splits: Mem-Noisy and Validation: 
Mem-Noisy is used to study the influence of MLP activations on memorizing noisy labels,
and Validation is used to study the influence of MLP activations on generalizing to clean labels.
We also study two settings: pre-memorization patching and random patching.
Similar to Appendix~\ref{app:probe_targets},
we use all instances from each data split as the clean prompts, 
and randomly sample another instance from the same split for each clean prompt as the corrupt prompt.
This results in four sets of experiments, as illustrated in Figure~\ref{subfig:layer_patching}. 

We follow the definition in Equation~\ref{eq:faithfulness} to compute the faithfulness score 
(we restate the metric here for clarity):
\begin{equation}
    \text{Faithfulness} = 
    \frac{\ell_{\text{patched}} - \ell_{\text{corrupt}}}
         {\ell_{\text{clean}} - \ell_{\text{corrupt}}}.
    \label{eq:faithfulness_pre_mem}
\end{equation}
Specifically, in this experiment, given a clean–corrupt prompt pair, 
we compute the logits for the correct answer token of the clean prompt under three conditions:

\begin{itemize}
    \item $\ell_{\text{patched}}$ is the mean logit from the patched run. 
    In the \emph{pre-memorization patching} setting, 
    we use the same clean prompt but replace the MLP activation 
    with that from the \emph{pre-memorization model}. 
    In the \emph{random patching} setting, 
    we use the same post-memorization model 
    but replace the MLP activation with that from the \emph{corrupt prompt}.

    \item $\ell_{\text{clean}}$ is the mean logit 
    for the correct clean answer token 
    when running the post-memorization model on the clean prompt.

    \item $\ell_{\text{corrupt}}$ is the mean logit 
    for the clean prompt's answer token when running the clean prompt, 
    but with \textbf{all MLP activations replaced} by those from the corrupt prompt. 
    (the post-memorization model is also used here). 
    We use this setup, instead of directly running the corrupt prompt, 
    because it provides a lower bound on the influence of MLP activations alone, 
    excluding changes in other parts of the model 
    (i.e., we do not consider the contributions of attention modules here).
\end{itemize}
We obtain all values for estimating faithfulness by averaging across 
all prompt pairs in the corresponding data split.

\section{Iterative Null-space Projection (INLP)}\label{app:inlp}
INLP is a popular method to remove linearly-encoded information 
from the hidden representations of a neural network~\citep{ravfogel2020null}. 
Specifically, it iteratively perform the following two steps:
(1) train a linear model to predict a target label from the hidden representations, 
e.g., the bridge entity in THR; and 
(2) project the hidden representations onto the null space of the linear model,
so that the linear model cannot predict the target label anymore.
This process is repeated multiple times, 
until it is no longer possible to train a linear model to predict the target label, 
i.e., the prediction accuracy is lower than a certain threshold $\epsilon$.
In our experiments, we set $\epsilon = 0.1$. 
In practice, we only need no more than three iterations to achieve this.

\section{Running Environment and AI usage}

\paragraph{Environment}
We use a single NVIDIA A100 GPU with 80GB memory for our experiments.
The main training and evaluation code is implemented in PyTorch~\citep{NEURIPS2019_bdbca288},
and Hugging Face Transformers~\citep{wolf-etal-2020-transformers}, using Python 3.12. 

\paragraph{Use of AI assistants}

Our code is implemented with the help of ChatGPT, Google Gemini, and GitHub Copilot. 
We mainly use these tools to assist data visualization after obtaining the results. 
Moreover, the person entity names and relation names, 
as well as the templates for verbalizing the triplets into sentences and QA pairs, 
are also constructed with the help of ChatGPT. 
We also used ChatGPT to assist paper writing, 
in particular to revise and improve the clarity of the text.

\clearpage
\section{Additional results}\label{app:additional_results}

\begin{figure}[ht]
    \centering
    \includegraphics[width=0.48\textwidth]{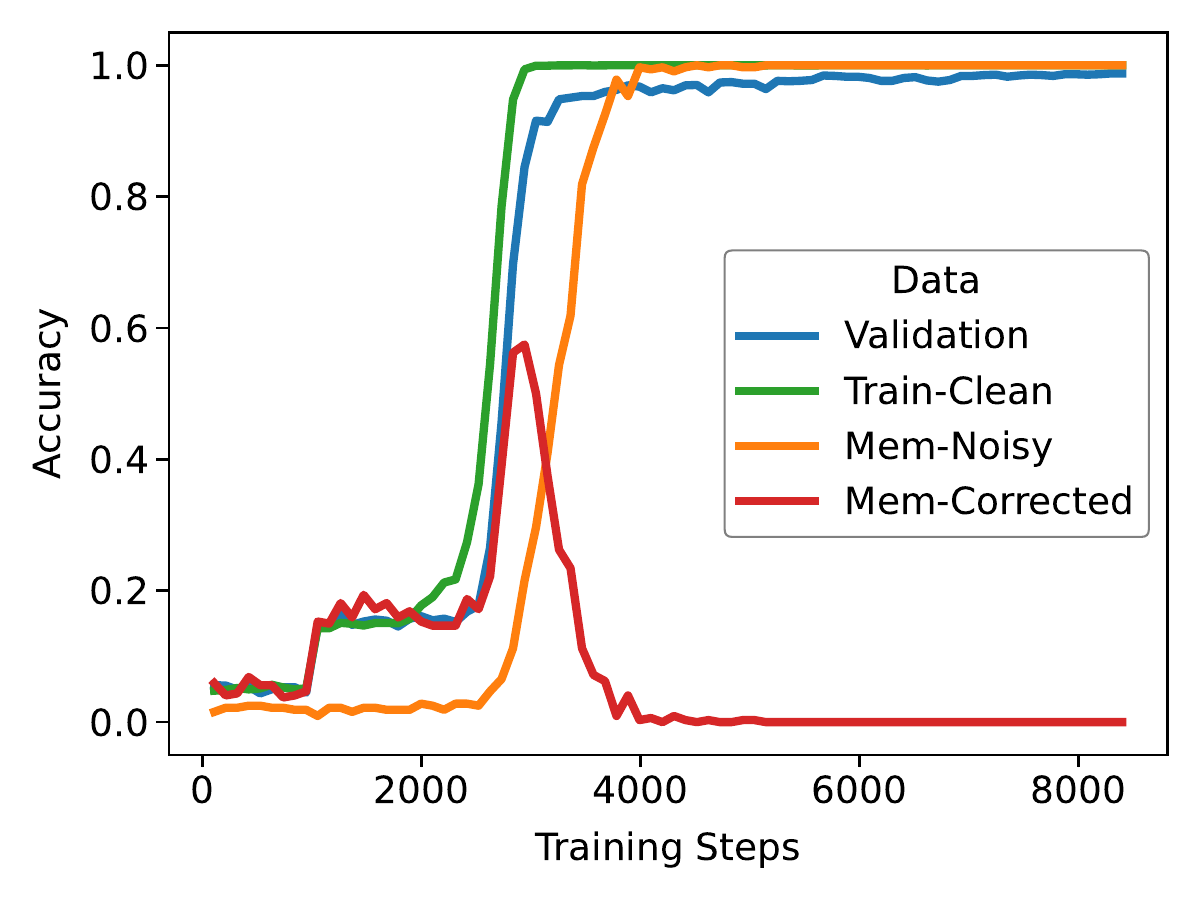}
    \caption{Learning dynamics of THR}
    \label{fig:learning_dynamics_thr}
\end{figure}

\paragraph{Learning dynamics of THR}
We show the learning dynamics of THR in Figure~\ref{fig:learning_dynamics_thr}, 
where we observe a similar trend as in FDA:
the model learns to produce the generalizable reasoning output first 
on training instances of noisy labels which the model has never seen 
(although the performance never reaches 100\%), 
but eventually memorizes the noisy labels.
Also, in the generalization stage, the model's performance on Mem-Corrected 
closely matches that of Validation. 

\subsection{Example outlier heuristics}

\begin{figure*}[ht!]
    \centering
    \includegraphics[width=0.95\textwidth]{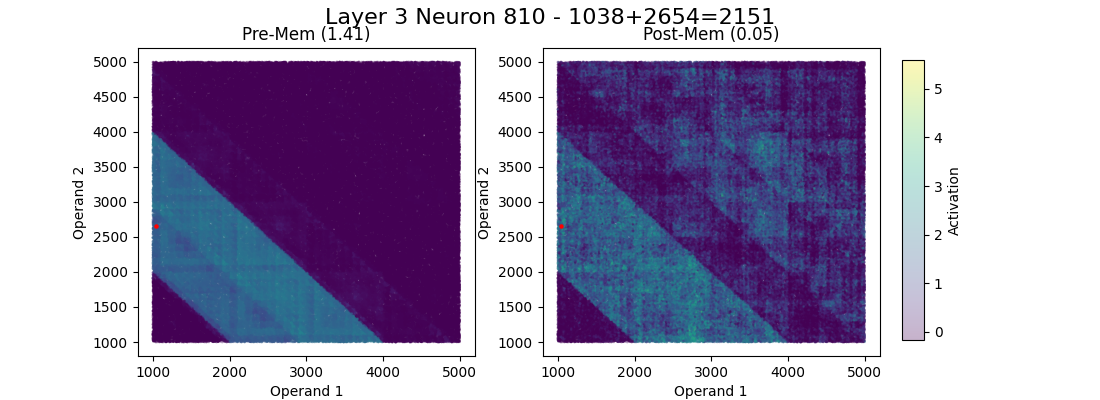}
    \includegraphics[width=0.95\textwidth]{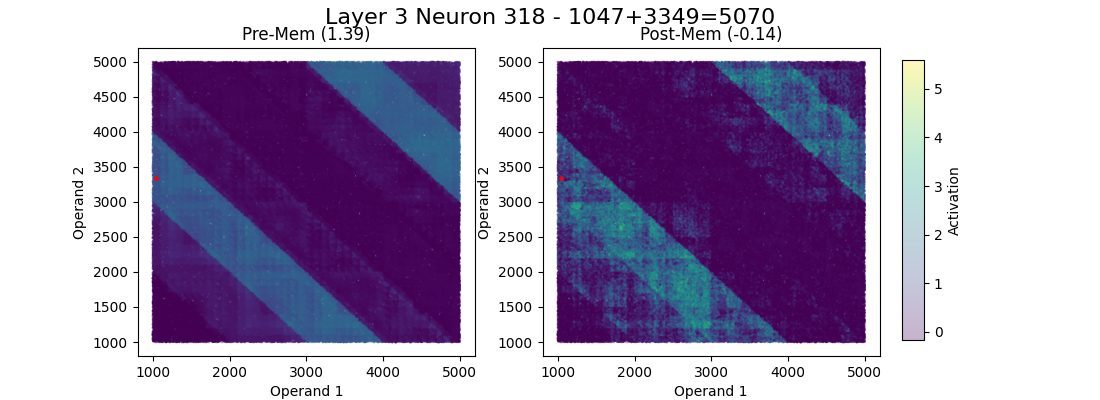}
    \includegraphics[width=0.95\textwidth]{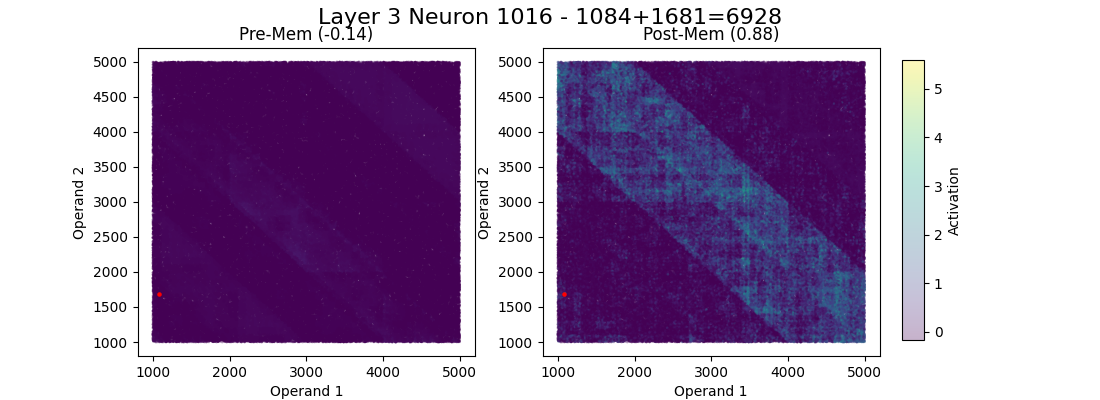}
    \includegraphics[width=0.95\textwidth]{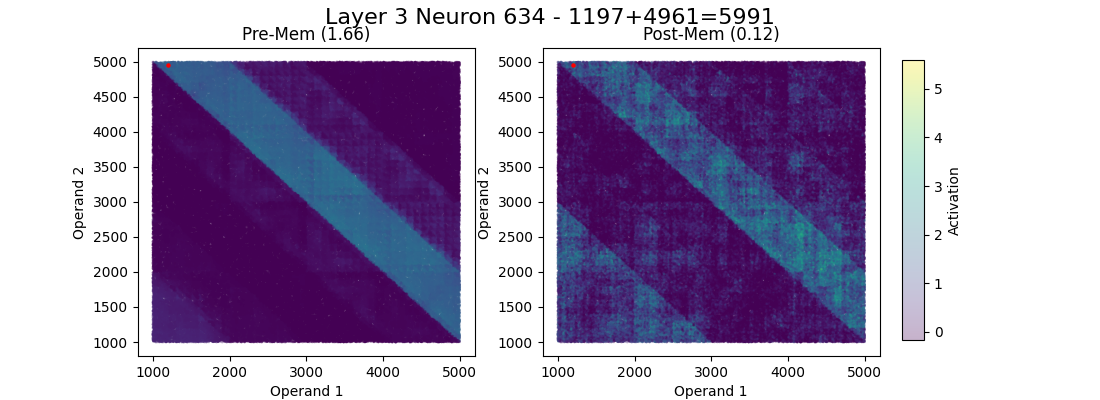}
    \caption{Examples of outlier heuristics: 
    Activation patterns of the most influential neuron for four different noisy training instances. 
    Each row shows the activation pattern for one instance. 
    The red dot indicates the position of the noisy training instance.}
    \label{fig:outlier_heuristics_examples}
\end{figure*}

We show more examples of outlier heuristics in Figure~\ref{fig:outlier_heuristics_examples}. 
Specifically, for each noisy training instance, 
we show the most influential neuron identified by 
the faithfulness drop by patching pre-memorization activation values. 
We consistently observe the outlier heuristics phenomenon. 

